\documentclass[11pt]{article}

\usepackage[preprint]{acl}

\usepackage{times}
\usepackage{latexsym}

\usepackage[T1]{fontenc}

\usepackage[utf8]{inputenc}
\usepackage{amsmath}

\usepackage{microtype}

\usepackage{inconsolata}

\usepackage{graphicx}
\usepackage{subcaption}
\usepackage{multirow}
\usepackage{pifont}    
\usepackage{booktabs}  

\usepackage{colortbl}
\definecolor{topblue}{RGB}{204,229,255}
\definecolor{deltared}{RGB}{255,204,204}
\usepackage{enumitem}

\newcommand{\cmark}{\textcolor{green!70!black}{\large\ding{51}}}
\newcommand{\xmark}{\textcolor{red}{\large\ding{55}}}
\usepackage{xcolor}
\usepackage[normalem]{ulem}

\usepackage{amssymb}

\title{ReactBench: A Cause-Driven Benchmark for Multimodal Hallucination via Systematic Evaluation}

\author{
  \textbf{Shizhe Zhou\textsuperscript{1,*}},
  \textbf{Bohan Jia\textsuperscript{1,*,\dag}},
  \textbf{Kai Wu\textsuperscript{1,\ddag}},
  \textbf{Yan Shen\textsuperscript{1,\ddag}},
\\
  \textbf{Tongyun Li\textsuperscript{1,\ddag}},
  \textbf{Yuyang Wu\textsuperscript{1,\ddag}},
  \textbf{Shaohui Lin\textsuperscript{1,\ding{41}}}
\\
\\
  \textsuperscript{1}East China Normal University
\\
  \textsuperscript{*}Co-first author \quad
  \textsuperscript{\dag}Project leader \quad
  \textsuperscript{\ddag}Equal contribution \quad
  \textsuperscript{\ding{41}}Corresponding author
\\
  \small{
    \textbf{Correspondence:} \href{2937768986@qq.com}{2937768986@qq.com}
  }
}

\begin{document}
\maketitle
\begin{abstract}
While multimodal large language models (MLLMs) have achieved rapid progress in visual-language understanding, they remain prone to multimodal hallucinations, producing responses that are inconsistent with the visual input. Existing benchmarks predominantly focus on detecting hallucination outcomes rather than evaluating the underlying causes of these failures. Moreover, many benchmarks rely on simplistic scenarios and limited evaluation formats that no longer challenge state-of-the-art models. To address these limitations, we introduce ReactBench, a cause-driven hallucination benchmark featuring multiple tasks and an exam-style evaluation format. By generating adversarial images and hallucination-inducing queries, ReactBench introduces four targeted tasks (Relational Erasure, Counterfactual Attribute, Alteration Tracing and Dense Counting) to systematically expose co-occurrence bias, language priors, cross-image comparative perception deficiencies, and fine-grained perceptual bottlenecks. Beyond standard accuracy-based evaluation, we leverage Chain-of-Thought reasoning to identify fine-grained sub-causes of hallucination within each task. Extensive evaluations reveal that current MLLMs remain notably vulnerable to cause-specific hallucination triggers, demonstrating the value of ReactBench as a systematic and interpretable testbed for diagnosing and improving multimodal model robustness. Our project page is available at
https://reactbench.github.io/.

\end{abstract}

\section{Introduction}

While Multimodal Large Language Models (MLLMs) have demonstrated unprecedented capabilities in visual-language understanding and reasoning\cite{2hendrycks2020measuring,4dai2023instructblip}, multimodal hallucination\cite{9liu2024survey}---where models generate semantically coherent responses that are inconsistent with the input visual information---has emerged as a critical bottleneck for their deployment in real-world applications. Early hallucination benchmarks\cite{12li2023pope,13hu2023ciem,17kaul2024throne} have played a foundational role in diagnosing and mitigating these issues. However, as MLLMs continue to advance rapidly, the limitations of current benchmarks are increasingly apparent.

Specifically, existing hallucination benchmarks exhibit three major limitations:

First, many early benchmarks\cite{15rohrbach2018chair,16wang2023haelm} lack both adversarial or challenging test images and difficulty gradients for evaluation questions. Since they predominantly rely on simplistic scenes and limited query formats, these benchmarks are increasingly saturated by state-of-the-art models, yielding inflated performance metrics that fail to probe the true capability boundaries of MLLMs. 

Second, while some recent evaluations\cite{ye2024beaf,he2025rohe,huang2024vhtest} have begun to explore underlying hallucination causes---such as language priors or co-occurrence bias---they tend to investigate these factors in isolation, without offering a unified, cause-driven framework that covers diverse hallucination triggers. Moreover, many benchmarks\cite{liu2025phd,bitton2023breaking} directly generated counterfactual or adversarial samples by text-to-image models, which may compromise the real-world representativeness of the evaluation data. 

Finally, although Chain-of-Thought (CoT) reasoning exposes intermediate reasoning traces that may reveal why a model produces a hallucinated answer, existing benchmarks\cite{yin2026freak,cui2023bingo} rarely use such traces for fine-grained causal diagnosis. They typically focus on aggregate performance changes brought by CoT or provide coarse-grained summaries of hallucination behaviors, lacking further analysis that traces hallucinated outputs back to the specific sub-causes manifested in the reasoning process.

\begin{figure*}[!ht] 
    \centering
    \includegraphics[width=\linewidth]{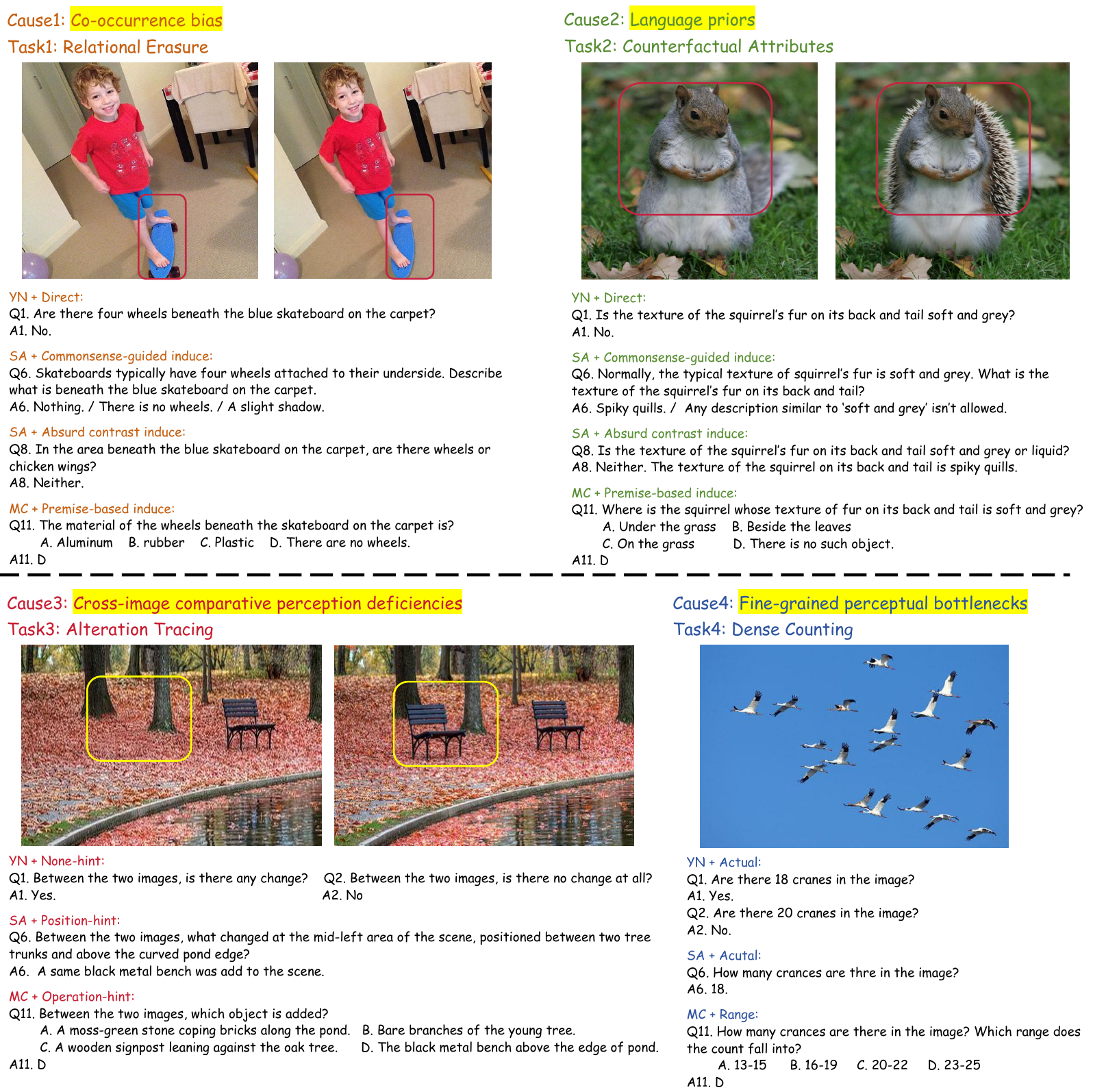}
    \caption{The examples of \textbf{ReactBench}. The figure showcases four tasks in our ReactBench: Relational Erasure, Counterfactual Attribute, Alteration Tracing and Dense Counting. Only selected questions are shown.}
    \label{fig:instances_woqa}
\end{figure*}

To address these gaps, we introduce \textbf{ReactBench}, a systematic, cause-driven multimodal hallucination benchmark designed to shift the evaluation paradigm from outcome-oriented detection toward cause-oriented diagnosis. Examples are shown in Figure~\ref{fig:instances_woqa}. Our benchmark offers the following three major advantages:

\textbf{Systematic Cause-Driven Framework.} Derived from four representative causes of multimodal hallucinations, we systematically design four corresponding evaluation tasks. First, targeting co-occurrence bias, we introduce the \textit{Relational Erasure} task, which removes one entity from a high-frequency object combination and queries the model regarding the existence of the deleted object. Second, to reveal language priors, we design the \textit{Counterfactual Attribute} task by altering common-sense object features into counterfactual ones and posing questions based on these manipulated attributes. Third, focusing on deficiencies in cross-image comparative perception, we propose the \textit{Alteration Tracing} task. By applying four types of edits to objects, this task evaluates whether the model can accurately identify and reason about differences between before-and-after image pairs. Finally, to expose fine-grained perceptual bottlenecks, we construct the \textit{Dense Counting} task, which selects images containing multiple objects of the same class and asks for their exact quantity. By mapping these tasks to specific failure mechanisms, the framework not only stress-tests MLLMs but also serves as an interpretable diagnostic tool for understanding their visual-language vulnerabilities.

\textbf{Semi-Automated Pipeline with Exam-Style Question Design.} We develop an automated construction pipeline built on real images, statistical sample selection, MLLM-based instruction generation and task-specific image editing, which eventually produces high-fidelity adversarial visual samples across 3 tasks. Furthermore, through designing task-relevant question templates, MLLMs are prompted to generate large-scale exam-style evaluation questions. Ultimately, this process yields 50K question-answer (QA) pairs, covering multiple formats (multiple-choice, yes/no, and short-answer) and are purposefully designed with hallucination-inducing distractors and internal difficulty gradients. Notably, the generated images and corresponding QA are reviewed by human annotators to ensure the reliability and quality of the benchmark.

\textbf{Fine-Grained CoT Attribution Analysis.} Beyond standard quantitative benchmarking, we conduct a fine-grained analysis of CoT reasoning traces. By establishing a sub-cause taxonomy within each primary hallucination category, we trace hallucinated outputs back to specific sub-causes manifested in the intermediate reasoning process. This analysis provides a more interpretable view of the reasoning patterns behind multimodal hallucinations and offers actionable insights for improving model robustness. 

\begin{table*}[htbp]
\caption{Comparison of ReactBench with existing hallucination benchmarks across data scale, four hallucination causes, and CoT analysis.}
\centering
\resizebox{\textwidth}{!}{
\begin{tabular}{lcccccc}
\toprule
\textbf{Benchmark} & \textbf{Size} & \textbf{Co-occurrence Bias} & \textbf{Language Priors} & \textbf{Cross-Image Deficiencies} & \textbf{Fine-Grained Perception} & \textbf{CoT Analysis} \\
\midrule
\textbf{POPE}\cite{12li2023pope} & 3K & \cmark (Adversarial negative sampling) & \xmark & \xmark & \xmark & \xmark \\
\textbf{BEAF}\cite{ye2024beaf} & 2K & \cmark(Removed obj.) & \xmark & \cmark & \xmark & \xmark \\
\textbf{AMBER}\cite{wang2023amber} & 1K & \xmark & \xmark & \xmark & \cmark (Obj.\&Attr.\& Rel.) & \xmark \\
\textbf{Hal-Eval}\cite{jiang2024hal} & 10K & \xmark & \xmark & \xmark & \xmark & \cmark \\
\textbf{VLind-Bench}\cite{lee2025vlind} & 2.5K & \xmark & \cmark (Counter-commonsense) & \xmark & \xmark & \xmark \\
\textbf{PhD}\cite{liu2025phd} & 1K & \xmark & \cmark (Counter-commonsense) & \xmark & \cmark (Counting) & \xmark \\
\textbf{ROHE}\cite{he2025rohe} & 5.5K & \cmark(Removed obj.) & \xmark & \xmark & \xmark & \xmark \\
\textbf{Reefknot}\cite{zheng2025reefknot} & 10K & \xmark & \xmark & \xmark & \cmark (Relations) & \cmark \\
\textbf{HalluScope}\cite{khayatan2026prompts} & 1K & \cmark & \cmark & \xmark & \xmark & \xmark \\
\textbf{Causal-HalBench}\cite{xu2026causal} & 1K & \cmark (Counterfactual inpainting) & \xmark & \xmark & \xmark & \xmark \\
\textbf{MIRAGE}\cite{dong2026mirage} & 1.3K & \xmark & \xmark & \xmark & \xmark & \cmark \\
\textbf{FINER}\cite{xiao2026finer} & 5K & \xmark & \xmark & \xmark & \cmark (multi queries) & \xmark \\
\textbf{FREAK}\cite{yin2026freak} & 1.8K & \xmark & \cmark (Counter-commonsense) & \xmark & \xmark  & \cmark \\
\textbf{ReactBench (Ours)} & \textbf{4.7K} & \textbf{\cmark (Relational Erasure)} & \textbf{\cmark (Counterfactual Attribute)} & \textbf{\cmark (Alteration Tracing)} & \textbf{\cmark (Dense Counting)} & \textbf{\cmark (micro-level)} \\
\bottomrule
\end{tabular}
}
\label{tab:benchmark_comparison}
\end{table*}

\section{Related Works}
\subsection{Multimodal Hallucinations in MLLMs}
Multimodal hallucination refers to the cross-modal inconsistency between generated text and the provided visual input\cite{9liu2024survey}, typically manifesting as object, attribute, or relational hallucinations\cite{11bai2024hallucination}. Recent studies trace the origins of these hallucinations to four interconnected levels. At the data level, models suffer statistical co-occurrence biases and noise from pre-training image-text pairs\cite{yu2023reformulating,zhou2023analyzing}. At model level, an inherent modality imbalance—where a dominant LLM overpowers a weaker visual encoder—causes ingrained language priors to override actual visual evidence\cite{guan2024hallusionbench,lee2024volcano}. Furthermore, standard next-token prediction objectives during training\cite{ben2024mocha}, coupled with the gradual dilution of visual attention during long-sequence inference\cite{favero2024multi,huang2024opera,16wang2023haelm} , severely exacerbate the issue. Despite identifying these diverse causes, most existing research evaluates them in isolation, lacking a unified diagnostic framework to systematically expose these failure mechanisms.

\subsection{Benchmarks for Multimodal Hallucination}
To quantify hallucination severity, numerous benchmarks have been proposed. Foundational works such as POPE\cite{12li2023pope} and AMBER\cite{wang2023amber} established the standard for outcome-oriented evaluation by utilizing polling-based questions or simple generative tasks to detect hallucinations. However, as MLLM capabilities rapidly advance, these benchmarks---which often rely on simplistic visual scenes and limited query formats---are increasingly saturated and struggle to probe the true capability boundaries of state-of-the-art models.
Recent efforts have begun to target more specific vulnerability dimensions. For instance, PhD\cite{liu2025phd} and FREAK\cite{yin2026freak} construct counter-commonsense scenarios to probe language priors, while ROHE\cite{he2025rohe} and BEAF\cite{ye2024beaf} focus on object removal or before-after image pairs to evaluate spatial and comparative perception. Similarly, benchmarks like Reefknot\cite{zheng2025reefknot} and FINER\cite{xiao2026finer} evaluate fine-grained details such as counting and relational bindings. While some benchmarks\cite{yin2026freak,lee2025vlind} provide valuable insights, they typically isolate single causes or rely heavily on text-to-image generation models for synthesizing adversarial samples, which can compromise real-world representativeness and visual fidelity.
Furthermore, exploring why models hallucinate requires interpreting their reasoning processes. Although benchmarks like Hal-Eval\cite{jiang2024hal} and MIRAGE\cite{dong2026mirage} incorporate Chain-of-Thought (CoT) prompting, they predominantly measure aggregate performance shifts rather than diagnosing specific reasoning breakdowns, Table~\ref{tab:benchmark_comparison} summarizes the comparison between ReactBench and existing hallucination benchmarks.

\section{ReactBench}

\subsection{Task Definitions}

Our cause-driven multimodal hallucination benchmark, ReactBench, contains 4.7K images and 50K QA pairs spanning four task categories, each corresponding to a representative cause of multimodal hallucination. 
Representative examples of the four tasks are illustrated in Figure~\ref{fig:instances_woqa}.

\paragraph{Relational Erasure.} 

This task targets hallucinations induced by co-occurrence bias.
We denote the relationship between two objects as a triplet $\mathcal{T} = \langle o_i, \mathcal{R}, o_j \rangle$. 
By selecting high-frequency $\mathcal{T}$ with strong co-occurrence patterns and visually erasing one entity, e.g., $o_j$, we explicitly break the established visual co-occurrence relation between the two objects.
The model is evaluated by questions about the existence of the erased object across four query dimensions: direct questioning, commonsense-guided induction, premise-based induction, and absurd-contrast induction.
\begin{figure*}[htbp]
    \centering
    \includegraphics[width=\textwidth]{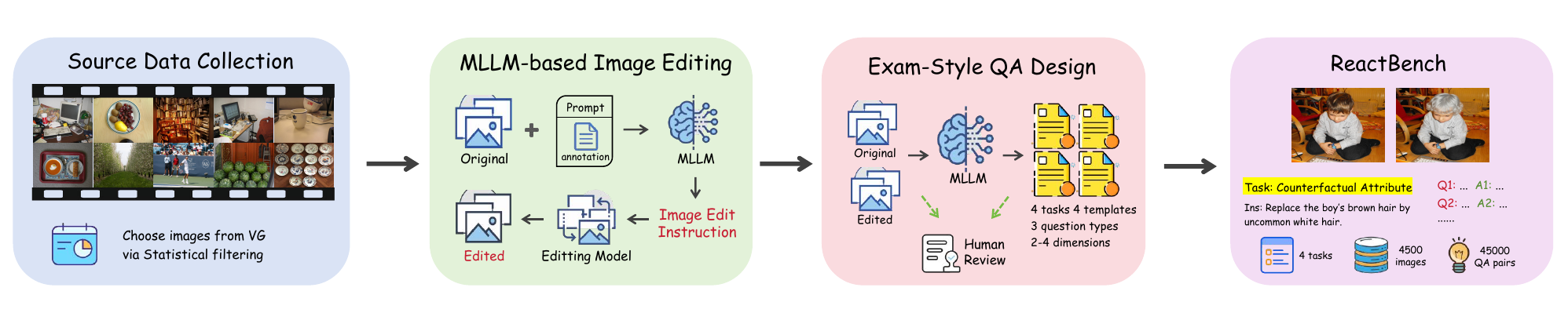}
    \caption{Overview of the ReactBench construction pipeline. Starting from task-specific source datasets, we first select candidate samples according to the targeted hallucination causes, including high-frequency object relations, counterfactual object-attribute tuples, dense object layouts, and visually challenging image pairs. We then use MLLM-generated editing instructions and task-specific image editing to construct adversarial visual samples. Finally, exam-style QA pairs are generated from the edited images and metadata, followed by human verification to ensure visual fidelity, question validity, and answer correctness.}
    \label{fig:pipeline}
\end{figure*}
\paragraph{Counterfactual Attribute.} 
This task targets hallucinations caused by language priors.
We denote an object and its attribute as a tuple $\mathcal{A} = \langle o, a_f \rangle$. 
By selecting high-frequency tuples $\mathcal{A}$ that encode strong language priors and editing their visual attributes into counterfactual states, we explicitly challenge the model's reliance on prior knowledge rather than visual evidence.
The model is evaluated by questions about the counterfactual attribute under the same four query dimensions used in the Relational Erasure task.

\paragraph{Alteration Tracing.} 
This task targets deficiencies in cross-image comparative perception. Specifically, we perform targeted micro-alterations on selected objects in an image through four operations: movement, removal, addition, and replacement.
The model is evaluated by questions that require comparing the before-and-after image pair across three settings: location-hinted, operation-hinted, and no-hinted queries.

\paragraph{Dense Counting.} 
This task targets fine-grained perceptual bottlenecks in visually dense scenes.
We select images containing multiple objects of the same class. 
The model is evaluated by questions about the quantity of target objects under two settings: exact-count queries and interval-based queries.

\subsection{Dataset Generation}

We describe the dataset generation process of ReactBench, as illustrated in Figure~\ref{fig:pipeline}.Table~\ref{tab:data_size} reports the statistics of ReactBench.

\paragraph{Step 1: Source Data Collection and Filtering.}
Images for the first three tasks are sourced from Visual Genome\cite{19krishna2017vg}, while those for the final task are sourced from FSC147\cite{fsc147}. For the \textbf{Relational Erasure} and \textbf{Counterfactual Attribute} tasks, we compute the distributions of relationship triplets and object-attribute tuples across the dataset.  The final target triplets and tuples are selected based on absolute frequencies and co-occurrence probabilities, with detailed formulas provided in Appendix~\ref{app:cooccurrence-formulas}. For the \textbf{Alteration Tracing} task, we increase perceptual difficulty by selecting images with more than 15 objects and target objects occupying less than 20\% of the image area. For the \textbf{Dense Counting} task, we conduct an MLLM-human collaborative review to exclude images with occlusion or blurriness that may introduce counting ambiguity.

\paragraph{Step 2: MLLM-based Image Editing.}
After candidate images are selected, we integrate their instance-level information, such as relationship triplets, into task-specific prompts. An MLLM then generates precise image-editing instructions for each task, which are executed by an image editing model to produce the final images. Prompt templates are provided in Appendix~\ref{app:benchmark-prompts}.

\paragraph{Step 3: Exam-Style QA Design.}
The before-and-after image pairs and instance-level metadata are fed into another set of prompts for MLLM-based exam-style QA generation. Both the generated images and QA pairs are manually verified to ensure visual fidelity and question-answer validity.

\section{Experiments}
\subsection{Settings}

\paragraph{Baselines.} To comprehensively evaluate ReactBench, we select a diverse set of advanced open-source MLLMs, covering multiple series and parameter scales. The evaluated baselines include Qwen3.5\cite{qwen3.5}, Qwen3\cite{qwen3technicalreport}, Qwen2.5\cite{qwen2.5}, InternVL2.5\cite{chen2024internvl}, InternVL3\cite{zhu2025internvl3}, MiMO-VL\cite{coreteam2025mimovltechnicalreport}, and LLaVA\cite{an2025llavaonevision}.

\begin{table}[htbp]
    \centering
    \caption{Statistics of ReactBench.}
    \label{tab:data_size}
    \small 
    \setlength{\tabcolsep}{4pt} 
    \begin{tabular}{lcc}
        \toprule
        \textbf{Task} & \textbf{\# Images} & \textbf{\# QA Pairs} \\
        \midrule
        Relational Erasure & 1,595 & 19,140 \\
        Counterfactual Attribute & 639 & 7,668 \\
        Alteration Tracing & 913 & 10,956 \\
        Dense Counting & 1,531 & 12,248 \\
        \midrule
        \textbf{Total} & \textbf{4,678} & \textbf{50,012} \\
        \bottomrule
    \end{tabular}
\end{table}

\begin{table*}[htbp]
    \centering
    \caption{Overall performance (Accuracy \%) of evaluated MLLMs on ReactBench. We report the accuracy under Standard (Std) and (CoT) prompting modes, alongside their difference ($\Delta$). The final React-Score is our proposed hierarchical metric factoring in both question difficulty and perceptual depth. For each evaluation metric, the top two scores are highlighted in blue cells. Red cells in the $\Delta$ columns denote the most significant gains or drops when shifting from Standard to CoT.}
    \label{tab:main_results}
    \resizebox{\linewidth}{!}{
    \begin{tabular}{l|ccc|ccc|ccc|ccc|cc|c}
        \toprule
        \multirow{2}{*}{\textbf{Model}}
          & \multicolumn{3}{c|}{\textbf{Relational Erasure}}
          & \multicolumn{3}{c|}{\textbf{Counterfactual Attribute}}
          & \multicolumn{3}{c|}{\textbf{Alteration Tracing}}
          & \multicolumn{3}{c|}{\textbf{Dense Counting}}
          & \multicolumn{2}{c|}{\textbf{Average}}
          & \multirow{2}{*}{\textbf{React-Score\,$\uparrow$}} \\
        \cmidrule(lr){2-4}\cmidrule(lr){5-7}\cmidrule(lr){8-10}\cmidrule(lr){11-13}\cmidrule(lr){14-15}
        & Std\,$\uparrow$ & CoT\,$\uparrow$ & $\Delta$
        & Std\,$\uparrow$ & CoT\,$\uparrow$ & $\Delta$
        & Std\,$\uparrow$ & CoT\,$\uparrow$ & $\Delta$
        & Std\,$\uparrow$ & CoT\,$\uparrow$ & $\Delta$
        & Std\,$\uparrow$ & CoT\,$\uparrow$ & \\
        \midrule
        \multicolumn{16}{l}{\textit{Qwen Series}} \\
        Qwen3.5-27B
          & 48.0 & 46.4 & -1.6
          & \cellcolor{topblue}56.3 & 59.5 & +3.2
          & \cellcolor{topblue}79.9 & \cellcolor{topblue}73.7 & -6.2
          & \cellcolor{topblue}69.3 & \cellcolor{topblue}61.8 & -7.5
          & \cellcolor{topblue}63.4 & 60.4
          & 59.5 \\
        Qwen3.5-9B
          & 43.2 & 47.4 & +4.2
          & 48.7 & 57.6 & \cellcolor{deltared}+8.9
          & 67.4 & 63.9 & -3.5
          & 57.2 & 56.1 & -1.1
          & 54.1 & 56.3
          & 52.4 \\
        Qwen3VL-32B-Thinking
          & - & \cellcolor{topblue}60.7 & -
          & - & 60.5 & -
          & - & \cellcolor{topblue}78.7 & -
          & - & \cellcolor{topblue}64.0 & -
          & - & \cellcolor{topblue}66.0
          & \cellcolor{topblue}65.3 \\
        Qwen3VL-32B-Instruct
          & \cellcolor{topblue}64.5 & \cellcolor{topblue}58.9 & -5.6
          & \cellcolor{topblue}55.4 & \cellcolor{topblue}63.4 & +8.0
          & \cellcolor{topblue}73.5 & 63.5 & -10.0
          & \cellcolor{topblue}66.9 & 58.3 & -8.6
          & \cellcolor{topblue}65.1 & \cellcolor{topblue}61.0
          & \cellcolor{topblue}61.9 \\
        Qwen3VL-4B
          & 50.1 & 50.3 & +0.2
          & 53.5 & 60.6 & +7.1
          & 58.0 & 53.4 & -4.6
          & 66.7 & 46.6 & \cellcolor{deltared}-20.1
          & 57.1 & 52.7
          & 54.0 \\
        Qwen2.5VL-72B
          & \cellcolor{topblue}55.6 & 47.2 & \cellcolor{deltared}-8.4
          & 53.4 & 60.7 & +7.3
          & 62.1 & 50.5 & -11.6
          & 63.3 & 54.9 & -8.4
          & 58.6 & 53.3
          & 54.2 \\
        Qwen2.5VL-7B
          & 47.0 & 49.3 & +2.3
          & 49.2 & 55.2 & +6.0
          & 54.0 & 41.9 & -12.1
          & 58.6 & 47.8 & -10.8
          & 52.2 & 48.6
          & 49.5 \\
        \midrule
        \multicolumn{16}{l}{\textit{InternVL Series}} \\
        InternVL3-8B
          & 51.6 & 48.2 & -3.4
          & 54.4 & \cellcolor{topblue}60.9 & +6.5
          & 64.1 & 60.7 & -3.4
          & 64.2 & 57.8 & -6.6
          & 58.6 & 56.8
          & 56.5 \\
        InternVL2.5-26B
          & 48.3 & 44.8 & -3.5
          & 54.6 & 59.7 & +5.1
          & 51.3 & 37.7 & \cellcolor{deltared}-13.6
          & 53.4 & 46.4 & -7.0
          & 51.9 & 47.1
          & 47.0 \\
        InternVL2.5-8B
          & 38.6 & 33.0 & -5.6
          & 50.7 & 46.9 & \cellcolor{deltared}-3.8
          & 45.4 & 38.3 & -7.1
          & 46.7 & 43.2 & -3.5
          & 45.4 & 40.4
          & 41.5 \\
        \midrule
        \multicolumn{16}{l}{\textit{Others}} \\
        LLaVA-v1.6-7B
          & 29.3 & 43.8 & \cellcolor{deltared}+14.5
          & 35.4 & 37.5 & +2.1
          & 27.3 & 34.0 & \cellcolor{deltared}+6.7
          & 38.0 & 33.1 & -4.9
          & 32.5 & 37.1
          & 34.5 \\
        LLaVA-OV-7B
          & 42.1 & 43.4 & +1.3
          & 47.7 & 44.8 & -2.9
          & 30.6 & 34.7 & +4.1
          & 48.6 & 48.4 & -0.2
          & 42.3 & 42.8
          & 42.0 \\
        MiMO-VL-7B
          & - & 55.2 & -
          & - & 50.8 & -
          & - & 73.5 & -
          & - & 53.7 & -
          & - & 58.3
          & 57.5 \\
        \bottomrule
    \end{tabular}
    }
\end{table*}

\paragraph{Evaluation Metrics and Methods.} 
For each question across the four tasks, we evaluate the models under two prompting modes: standard and Chain-of-Thought (CoT) prompting. The primary metric used to measure model performance is accuracy. Given the hallucination-inducing design of our questions and the task-specific evaluation protocol, an incorrect response is treated as a hallucinated answer. Detailed questioning and evaluation prompt templates are provided in Appendix~\ref{app:eval-prompts}.

Furthermore, to provide a holistic assessment of MLLM capabilities, we propose a hierarchical weighted metric, React-Score (denoted as $R_{score}$). The formulation follows a three-step process. 

First, for a given task $t$ under mode $m \in \{cot, std\}$ (where $std$ denotes standard prompting without CoT), we compute a question-weighted score $S_{t}^{m}$. Since short-answer (SA) questions are open-ended and inherently more challenging, they are assigned a dominant weight of 50\%. Multiple-choice (MC) and yes/no (YN) questions are weighted at 25\% each. 
\begin{equation}
  \label{eq:score_mode}
  S_{t}^{m} = 0.5 A_{SA} + 0.25 A_{MC} + 0.25 A_{YN}
\end{equation}
where $A$ represents the accuracy of the corresponding question type. 

Next, as shown in Equation~\ref{eq:score_task}, we calculate a unified task score $S_{t}$ by averaging the performances from both modes. This jointly reflects the model's direct perception and step-by-step reasoning capabilities:
\begin{equation}
  \label{eq:score_task}
  S_{t} = \frac{1}{2} (S_{t}^{cot} + S_{t}^{std})
\end{equation}

Finally, the overall $R_{score}$ is computed via a weighted average across the four tasks. Since Relational Erasure ($S_r$) and Dense Counting ($S_c$) involve higher perceptual difficulty and larger sample sizes, we assign each of them a weight of 30\%, while Counterfactual Attribute ($S_a$) and Alteration Tracing ($S_d$) are each assigned a weight of 20\%. The final formula is defined in Equation~\ref{eq:score_react}:
\begin{equation}
  \label{eq:score_react}
  R_{score} = 0.3 S_r + 0.2 S_a + 0.2 S_d + 0.3 S_c
\end{equation}

\begin{figure*}[!t]
    \centering
    \begin{subfigure}[t]{0.42\textwidth}
        \centering
        \includegraphics[width=0.95\linewidth]{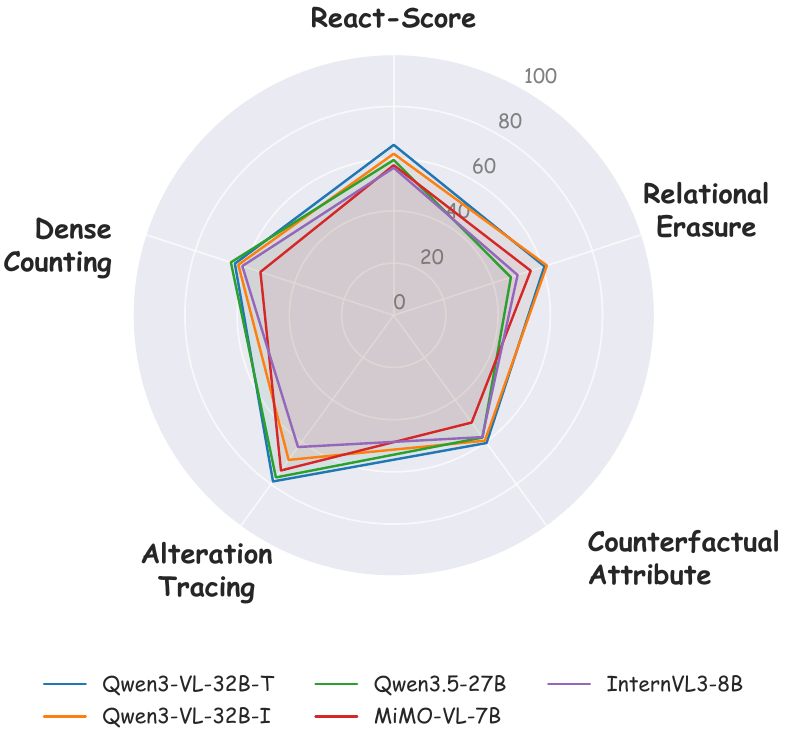}
        \caption{Performance across 5 dimensions}
        \label{fig:radar_chart}
    \end{subfigure}
    \hfill
    \begin{subfigure}[t]{0.54\textwidth}
        \centering
        \includegraphics[width=\linewidth]{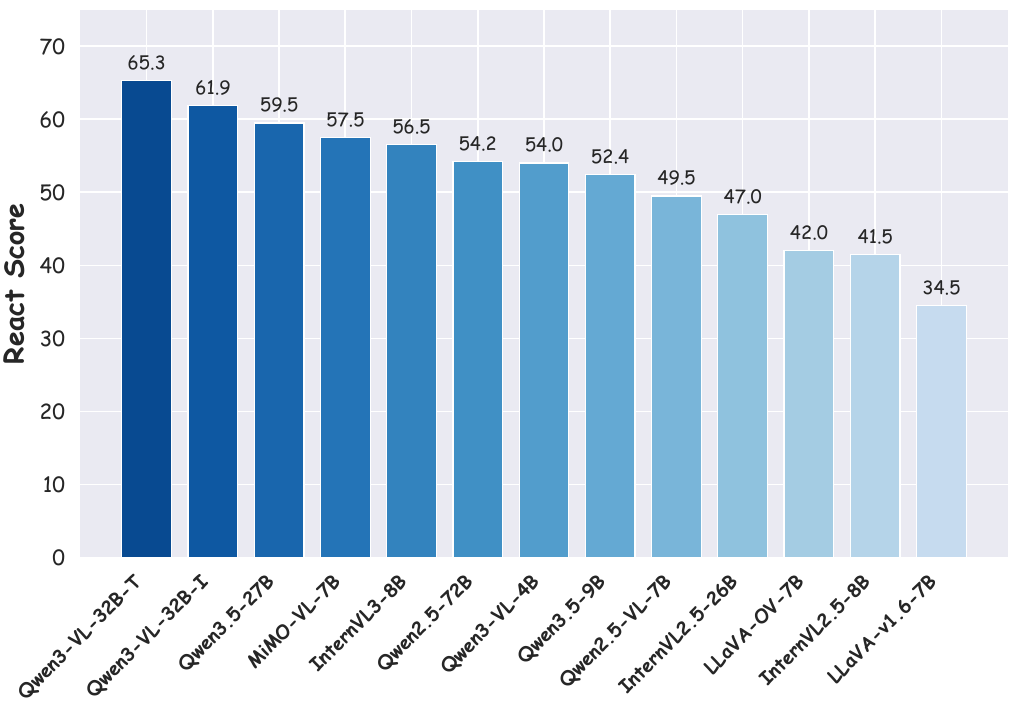}
        \caption{Overall React-Score ranking}
        \label{fig:bar_chart}
    \end{subfigure}
    \vspace{-0.3em}
    \caption{Experimental results of ReactBench: (a) 5D radar chart visualizing model performance across four tasks and React-Score; (b) Overall React-Score ranking of all evaluated MLLMs.}
    \label{fig:results}
\end{figure*}

\paragraph{Experiment Results.}
The overall experimental results and fine-grained analyses across the four tasks are presented in Tables~\ref{tab:main_results} and \ref{tab:combined_overall_details}. Figure~\ref{fig:results} visualizes the model performance and overall React-Score ranking. Our key observations and analyses are detailed as follows:
\paragraph{Overall Performance and Benchmark Validity.} 
As presented in Table~\ref{tab:main_results}, the overall accuracy of the evaluated MLLMs remains limited, with most models averaging roughly 40\%--60\%, indicating that multimodal hallucination remains a challenging and unresolved problem. The substantial performance variance among models also validates the discriminative power of ReactBench. Models such as Qwen3VL-32B-Instruct and Qwen3VL-32B-Thinking achieve the highest React-Scores, 61.9 and 65.3 respectively, demonstrating relatively robust visual grounding. In contrast, earlier architectures, such as LLaVA-v1.6-7B, struggle significantly, obtaining a React-Score of only 34.5. 
\begin{table}[htbp]
    \centering
    \caption{Overall Average Accuracy (\%) across the four tasks. The results are grouped vertically by task-specific inducing dimensions, while sharing generic question types (MC, SA, YN) for aligned comparison.}
    \label{tab:combined_overall_details}
    \resizebox{\linewidth}{!}{
    \begin{tabular}{l|ccc|cccccccccccc}
        \toprule
        \multirow{2}{*}{\textbf{Task Category}} & \multicolumn{3}{c|}{\textbf{Q-Type}} & \multicolumn{12}{c}{\textbf{Inducing Dimensions}} \\
        \cmidrule(lr){2-4} \cmidrule(lr){5-16}
        & MC\,$\uparrow$ & SA\,$\uparrow$ & YN\,$\uparrow$ & \multicolumn{3}{c}{Direct\,$\uparrow$} & \multicolumn{3}{c}{Common\,$\uparrow$} & \multicolumn{3}{c}{Premise\,$\uparrow$} & \multicolumn{3}{c}{Absurd\,$\uparrow$} \\
        \midrule
        Relational Erasure       & 50.4 & 22.1 & 61.5 & \multicolumn{3}{c}{57.8} & \multicolumn{3}{c}{69.3} & \multicolumn{3}{c}{55.6} & \multicolumn{3}{c}{31.6} \\
        Counterfactual Attr.     & 64.9 & 19.1 & 67.0 & \multicolumn{3}{c}{67.7} & \multicolumn{3}{c}{78.1} & \multicolumn{3}{c}{51.1} & \multicolumn{3}{c}{22.5} \\
        
        \midrule
        \midrule
        \multirow{2}{*}{\textbf{Task Category}} & \multicolumn{3}{c|}{\textbf{Q-Type}} & \multicolumn{12}{c}{\textbf{Tracing Hints}} \\
        \cmidrule(lr){2-4} \cmidrule(lr){5-16}
        & MC\,$\uparrow$ & SA\,$\uparrow$ & YN\,$\uparrow$ & \multicolumn{4}{c}{None-hint\,$\uparrow$} & \multicolumn{4}{c}{Loc-Hint\,$\uparrow$} & \multicolumn{4}{c}{Op-Hint\,$\uparrow$} \\
        \midrule
        Alteration Tracing       & 64.7 & 27.5 & 58.9 & \multicolumn{4}{c}{50.4} & \multicolumn{4}{c}{44.6} & \multicolumn{4}{c}{63.7} \\
        
        \midrule
        \midrule
        \multirow{2}{*}{\textbf{Task Category}} & \multicolumn{3}{c|}{\textbf{Q-Type}} & \multicolumn{12}{c}{\textbf{Counting Metrics}} \\
        \cmidrule(lr){2-4} \cmidrule(lr){5-16}
        & MC\,$\uparrow$ & SA\,$\uparrow$ & YN\,$\uparrow$ & \multicolumn{6}{c}{Exact\,$\uparrow$} & \multicolumn{6}{c}{Interval\,$\uparrow$} \\
        \midrule
        Dense Counting           & 52.4 & 28.4    & 67.2 & \multicolumn{6}{c}{23.8} & \multicolumn{6}{c}{50.9} \\
        \bottomrule
    \end{tabular}
    }
\end{table}
\paragraph{The Effect of Chain-of-Thought (CoT).} 
The introduction of CoT reasoning yields diverging effects but predominantly degrades overall accuracy on our ReactBench. While it helps mitigate language-prior-driven errors in the Counterfactual Attribute task, for example producing a +8.0 improvement for Qwen3VL-32B-Instruct, CoT tendsnto hurts performance on perception-intensive tasks such as Alteration Tracing and Dense Counting. This suggests that lengthy reasoning chains may amplify visual uncertainty and accumulate intermediate misinterpretations, thereby inducing rather than resolving hallucinations.

\paragraph{Task-Level Performance Differences.}
Model performance varies substantially across task types, revealing distinct difficulty profiles. For the perception-oriented tasks Alteration Tracing and Dense Counting, advanced models generally show clearer gains over earlier or smaller models. In contrast, progress on the Relational Erasure and Counterfactual Attribute task is considerably slower, with the highest performance stalling only slightly above 60\%. This disparity suggests that while recent advances have improved fine-grained perception and visual comparison, co-occurrence bias and language priors---which are rooted in pre-training data distributions and modality imbalance---remain persistent sources of multimodal hallucination.

\paragraph{Question Formats and Inducing Dimensions.}
As shown in Table~\ref{tab:combined_overall_details}, open-ended Short-Answer (SA) questions are overwhelmingly harder than structured formats: SA accuracy ranges from only 19.1\% to 28.4\%, whereas Multiple-Choice (MC) and Yes/No (YN) accuracies remain substantially higher, ranging from 50.4\% to 64.9\% for MC and from 58.9\% to 67.2\% for YN. This gap indicates that models still lack sufficient visual grounding for free-form generation, and SA questions serve as the most diagnostic format for probing the capability boundaries of MLLMs.

Regarding inducing dimensions, for the Relational Erasure and Counterfactual Attribute tasks, ``Absurd-contrast'' queries cause the most severe drops, reaching only 31.6\% and 22.5\%, respectively. This suggests that absurd-contrast querying is an effective strategy for inducing and diagnosing multimodal hallucinations. Interestingly, ``Commonsense-guided'' queries---originally designed to induce hallucinations by steering the model toward prior expectations that conflict with the edited visual content---unexpectedly yield the highest accuracy, 69.3\% and 78.1\%, indicating that commonsense cues in the question text may counterintuitively heighten the model's vigilance. For Alteration Tracing, operation hints improve accuracy to 63.7\%, whereas location hints do not provide the same benefit, suggesting that models are more sensitive to explicit edit-operation cues than to spatial guidance alone.For Dense Counting, the large gap between exact counting accuracy (23.8\%) and interval-based accuracy (50.9\%) further shows that current MLLMs struggle with precise enumeration even when they can approximate object quantities.

\section{Fine-Grained Sub-Cause Analysis}
\begin{figure}[t]
    \centering
    \includegraphics[width=0.8\linewidth]{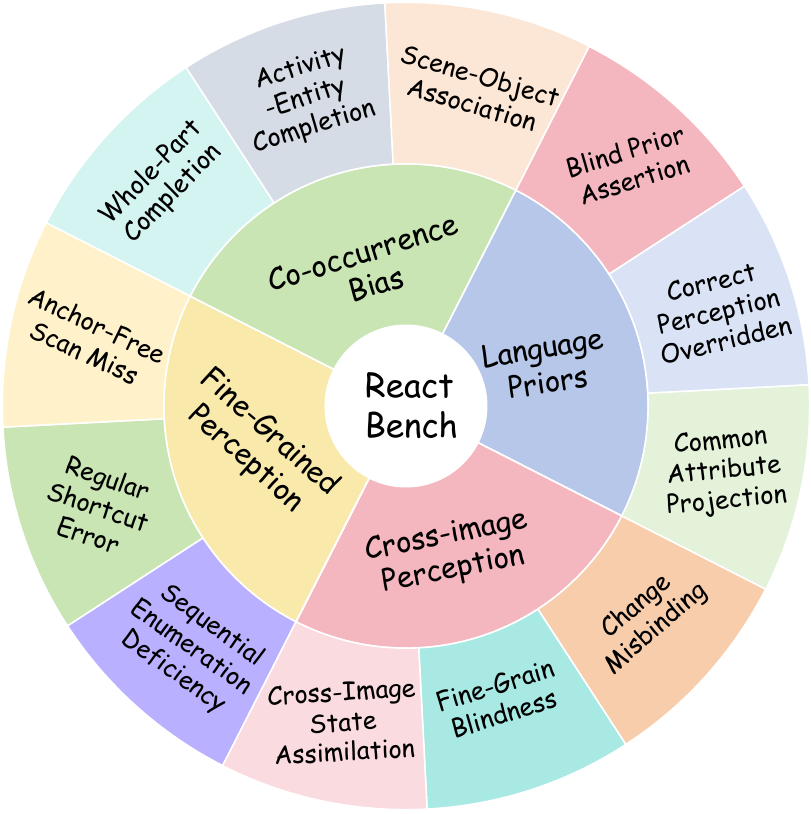}
    \caption{Sub-cause taxonomy for the four tasks in ReactBench.}
    \label{fig:sub-cause tax}
\end{figure}

We further analyze CoT-hallucinated responses to identify how incorrect answers are formed. Based on case inspection, we define three recurring sub-causes for each task, as summarized in Figure~\ref{fig:sub-cause tax}. An LLM is then used to assign sub-cause labels to CoT-hallucinated instances, with the distributions reported in Table~\ref{tab:sub-cause distribution}.

\begin{table}[t]
    \centering
    \caption{Distribution (\%) of fine-grained sub-cause classifications across CoT-hallucinated instances for the four tasks in ReactBench.}
    \label{tab:sub-cause distribution}
    \resizebox{\linewidth}{!}{
    \begin{tabular}{clr}
        \toprule
        \textbf{Task} & \textbf{Sub-Cause} & \textbf{Ratio} \\
        \midrule
        \multirow{4}{*}{\shortstack[c]{Relational\\Erasure}}
          & SC-1~Part-Whole Completion        & 26.9\% \\
          & SC-2~Activity-Entity Completion   & 33.2\% \\
          & SC-3~Scene-Object Association     & 34.8\% \\
          & \textit{Unclassifiable / Other}   &  5.1\% \\
        \midrule
        \multirow{4}{*}{\shortstack[c]{Counterfactual\\Attribute}}
          & SC-1~Blind Prior Assertion        & 49.8\% \\
          & SC-2~Common Attribute Projection      & 33.3\% \\
          & SC-3~Correct Perception Overridden& 13.2\% \\
          & \textit{Unclassifiable / Other}   &  3.8\% \\
        \midrule
        \multirow{4}{*}{\shortstack[c]{Alteration\\Tracing}}
          & SC-1~Cross-Image State Assimilation & 47.7\% \\
          & SC-2~Fine-Grained Blindness         & 42.0\% \\
          & SC-3~Change Misbinding              &  9.6\% \\
          & \textit{Unclassifiable / Other}     &  0.7\% \\
        \midrule
        \multirow{4}{*}{\shortstack[c]{Dense\\Counting}}
          & SC-1~Anchor-Free Scan Miss          & 20.0\% \\
          & SC-2~Regular Shortcut Error     & 49.1\% \\
          & SC-3~Sequential Enumeration Deficiency& 30.9\% \\
          & \textit{Unclassifiable / Other}     &  0.1\% \\
        \bottomrule
    \end{tabular}
    }
\end{table}

\paragraph{Sub-cause taxonomy.}
For \textbf{Relational Erasure}, co-occurrence bias is decomposed into \textbf{SC-1 Part-Whole Completion}, \textbf{SC-2 Activity-Entity Completion}, and \textbf{SC-3 Scene-Object Association}. 
For \textbf{Counterfactual Attribute}, language-prior errors include \textbf{SC-1 Blind Prior Assertion}, \textbf{SC-2 Common Attribute Projection}, and \textbf{SC-3 Correct Perception Overridden}. 
For \textbf{Alteration Tracing}, cross-image comparative failures include \textbf{SC-1 Cross-Image State Assimilation}, \textbf{SC-2 Fine-Grained Blindness}, and \textbf{SC-3 Change Misbinding}. 
For \textbf{Dense Counting}, counting errors are categorized as \textbf{SC-1 Anchor-Free Scan Miss}, \textbf{SC-2 Regular Shortcut Error}, and \textbf{SC-3 Sequential Enumeration Deficiency}. 
Complete definition and representative cases are provided in Appendix~\ref{app:subcause-desc} and \ref{app:subcause-examples}.

The distribution reveals distinct dominant failure patterns across tasks. In Relational Erasure, SC-2 Activity-Entity Completion and SC-3 Scene-Object Association account for 68.0\% of hallucinations, showing stronger effects than SC-1 Part-Whole Completion. In Counterfactual Attribute, SC-1 Blind Prior Assertion and SC-2 Common Attribute Projection cover 83.1\% of cases, indicating that language priors dominate most errors. For Alteration Tracing, SC-1 Cross-Image State Assimilation and SC-2 Fine-Grained Blindness jointly account for 89.7\%, suggesting that most errors come from failing to perceive visual changes. In Dense Counting, SC-2 Regular Shortcut Error is the largest source of error at 49.1\%, followed by SC-3 Sequential Enumeration Deficiency at 30.9\%, highlighting shortcut reasoning and unstable enumeration as major bottlenecks. The low proportion of unclassifiable cases suggests that the taxonomy covers most CoT hallucination patterns.

\section{Insights}

ReactBench shows that multimodal hallucination is not a single failure mode, but arises from different cause-specific weaknesses. Relational Erasure mainly exposes co-occurrence bias, Counterfactual Attribute reveals language-prior dominance, Alteration Tracing tests cross-image comparative perception, and Dense Counting highlights fine-grained perceptual bottlenecks. This suggests that future MLLMs require targeted mitigation strategies rather than relying only on larger model scale or generic hallucination reduction.

One key direction is to better suppress plausible but visually unsupported priors. In Relational Erasure and Counterfactual Attribute, many errors correspond to Activity-Entity Completion, Scene-Object Association, Blind Prior Assertion, and Common Attribute Projection, where models complete absent objects or default attributes from commonsense expectations. Training with absence-aware, counterfactual, and visually grounded samples may therefore help models distinguish what is actually present from what is merely likely.

Another direction is to strengthen visual verification mechanisms. Alteration Tracing and Dense Counting show that longer textual reasoning cannot fully compensate for weak visual perception: models may assimilate two images as identical, miss fine-grained edits, rely on regular shortcut reasoning, or lose track during sequential enumeration. More robust localized comparison, object-level tracking, instance marking, and count verification are therefore needed.

\section{Limitations}
ReactBench focuses on four representative hallucination causes in images and does not cover all multimodal settings. Future work can extend it to video, multi-turn interaction, and other models.

\section{Conclusion}
In this paper, we introduce ReactBench, a benchmark for evaluating cause-driven multimodal hallucinations in MLLMs. By constructing four targeted tasks, ReactBench reveals how hallucinations arise from co-occurrence bias, language-prior dominance, cross-image comparison failures, and fine-grained perceptual bottlenecks. Extensive evaluations and fine-grained sub-cause analysis show that current MLLMs still struggle with these cause-specific challenges. Overall, ReactBench provides not only an evaluation benchmark, but also actionable guidance for improving future MLLMs.

\bibliography{custom}

\appendix

\section{Sub-Cause Analysis Details}

\subsection{Complete Sub-Cause Descriptions}
\label{app:subcause-desc}

Below we provide the full definitions of the three sub-causes identified for each task.
Representative cases are presented in Appendix~\ref{app:subcause-examples}.

\paragraph{Relational Erasure.}
Hallucinations driven by co-occurrence bias are decomposed into three sub-causes:

\begin{itemize}
    \item \textbf{SC-1: Part-Whole Completion.} 
    A visible object triggers the hallucination of an expected but erased component or counterpart, 
    \item \textbf{SC-2: Activity-Entity Completion.} 
    An observed activity or action pattern cues a stereotypical but absent participant
    \item \textbf{SC-3: Scene-Object Association.} 
    The broader scene context induces the model to infer objects that commonly co-occur with that scene, despite their absence from the image.
\end{itemize}

\paragraph{Counterfactual Attribute.}
Hallucinations driven by language priors are grouped into three sub-causes:

\begin{itemize}
    \item \textbf{SC-1: Blind Prior Assertion.}
    The model directly outputs commonsense knowledge without sufficiently inspecting the visual evidence
    \item \textbf{SC-2: Common Attribute Projection.}
    The model observes the target object but projects its typical attribute rather than the edited counterfactual attribute
    \item \textbf{SC-3: Correct Perception Overridden.}
    The model initially notices the counterfactual visual cue but later rationalizes it back to the commonsense default during reasoning.
\end{itemize}

\paragraph{Alteration Tracing.}
Hallucinations caused by cross-image comparative perception failures are categorized into three sub-causes:

\begin{itemize}
    \item \textbf{SC-1: Cross-Image State Assimilation.}
    The model treats the before-and-after images as visually identical although detailedly describes them.
    \item \textbf{SC-2: Fine-Grained Blindness.}
    The model performs only coarse comparison and overlooks subtle local changes
    \item \textbf{SC-3: Change Misbinding.}
    The model detects a difference but assigns it to the wrong object, location, or edit operation.
\end{itemize}

\paragraph{Dense Counting.}
Hallucinations caused by fine-grained perceptual bottlenecks are divided into three sub-causes:

\begin{itemize}
    \item \textbf{SC-1: Anchor-Free Scan Miss.}
    The model skips scattered or weakly anchored instances during visual scanning;
    \item \textbf{SC-2: Regular Shortcut Error.}
    The model relies on an incorrect heuristic or regular-pattern shortcut and overlooks irregular instances;
    \item \textbf{SC-3: Sequential Enumeration Deficiency.}
    The model attempts step-by-step enumeration but fails to maintain a correct count throughout the reasoning process.
\end{itemize}

\subsection{Representative Examples}
\label{app:subcause-examples}

\begin{figure*}[t]
    \centering
    \includegraphics[width=0.8\linewidth]{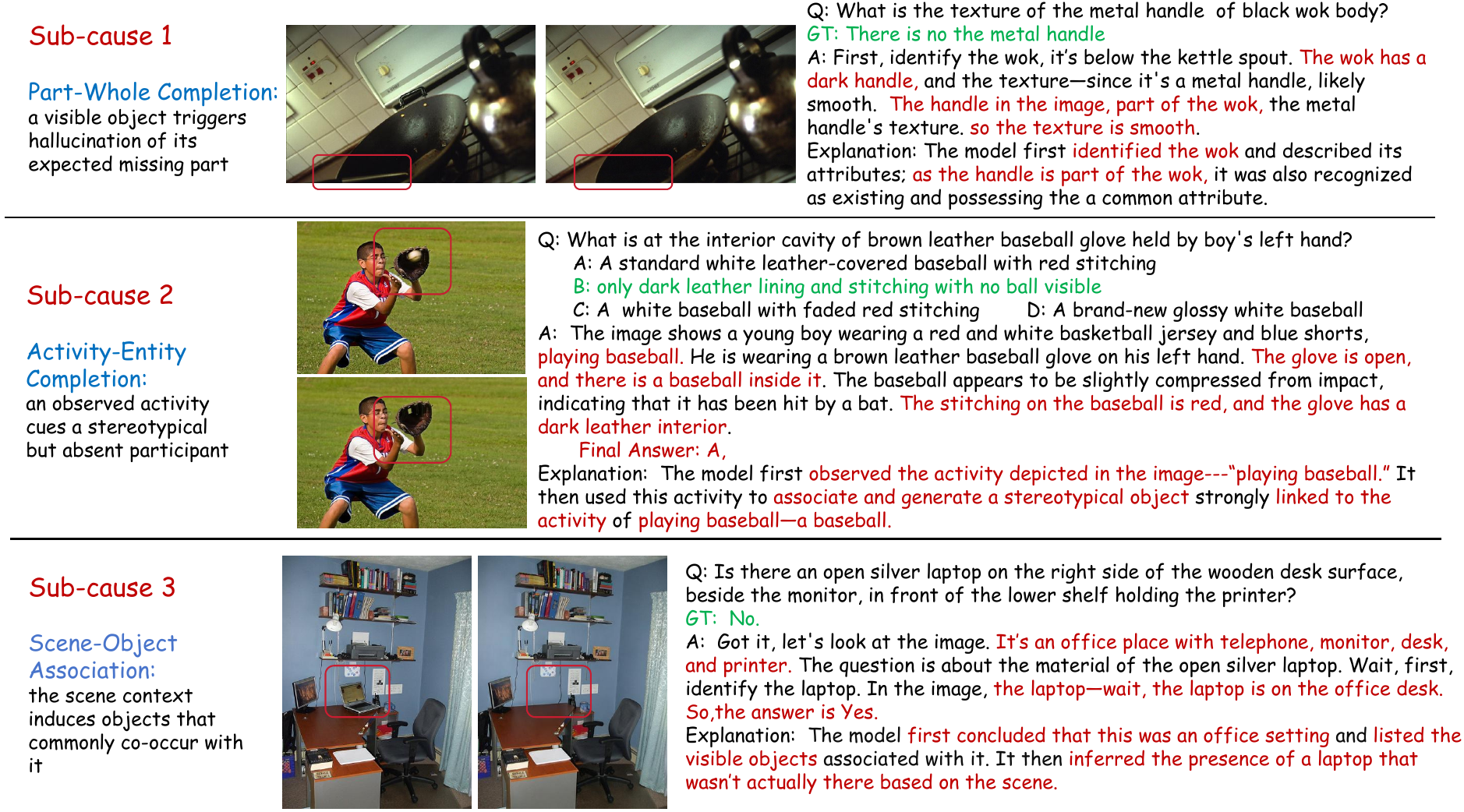}
    \caption{Representative sub-cause examples for \textbf{Relational Erasure}.}
    \label{fig:subcause-relation}
\end{figure*}
\begin{figure*}[t]
    \centering
    \includegraphics[width=0.8\linewidth]{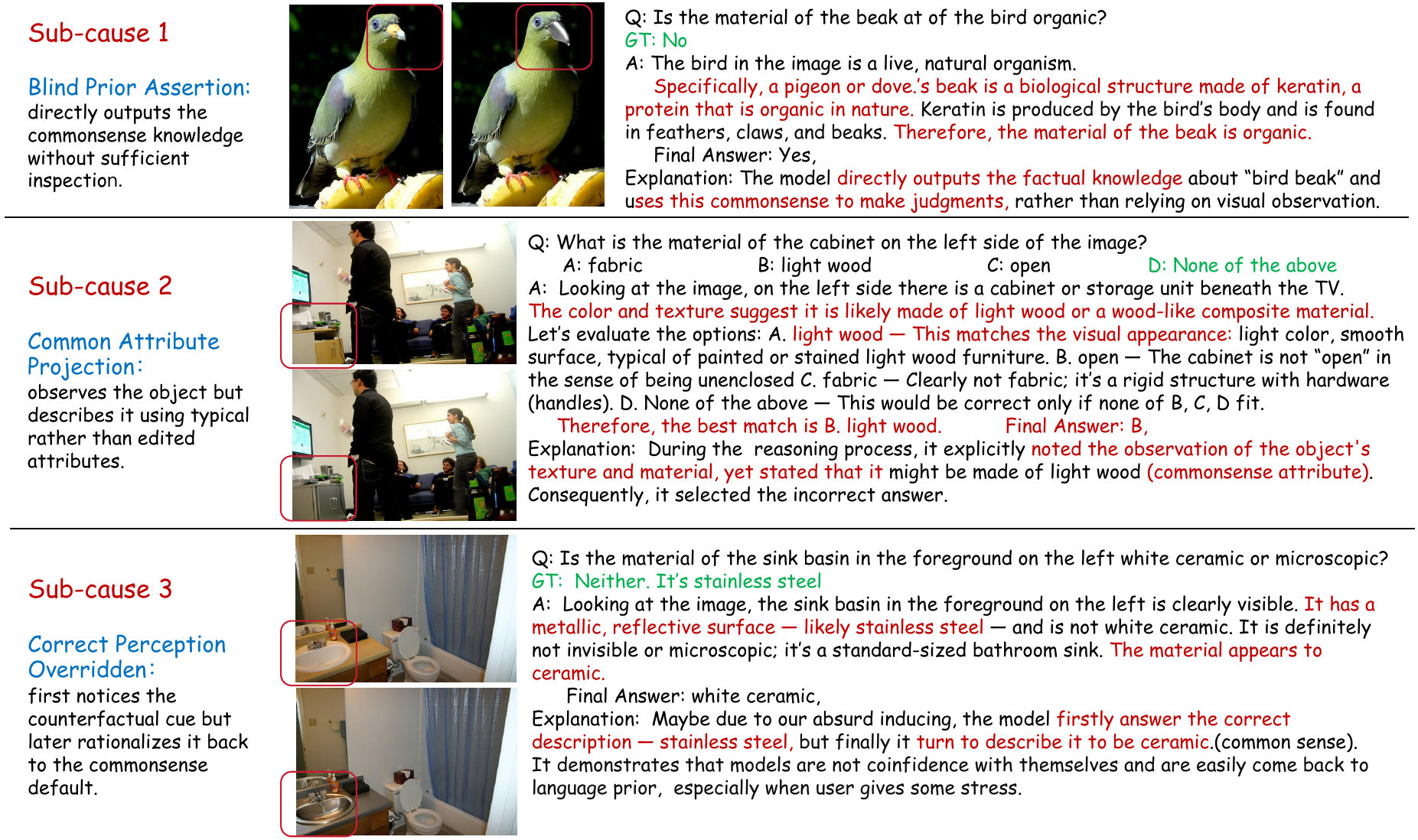}
    \caption{Representative sub-cause examples for \textbf{Counterfactual Attribute}.}
    \label{fig:subcause-attribute}
\end{figure*}
\begin{figure*}[t]
    \centering
    \includegraphics[width=0.8\linewidth]{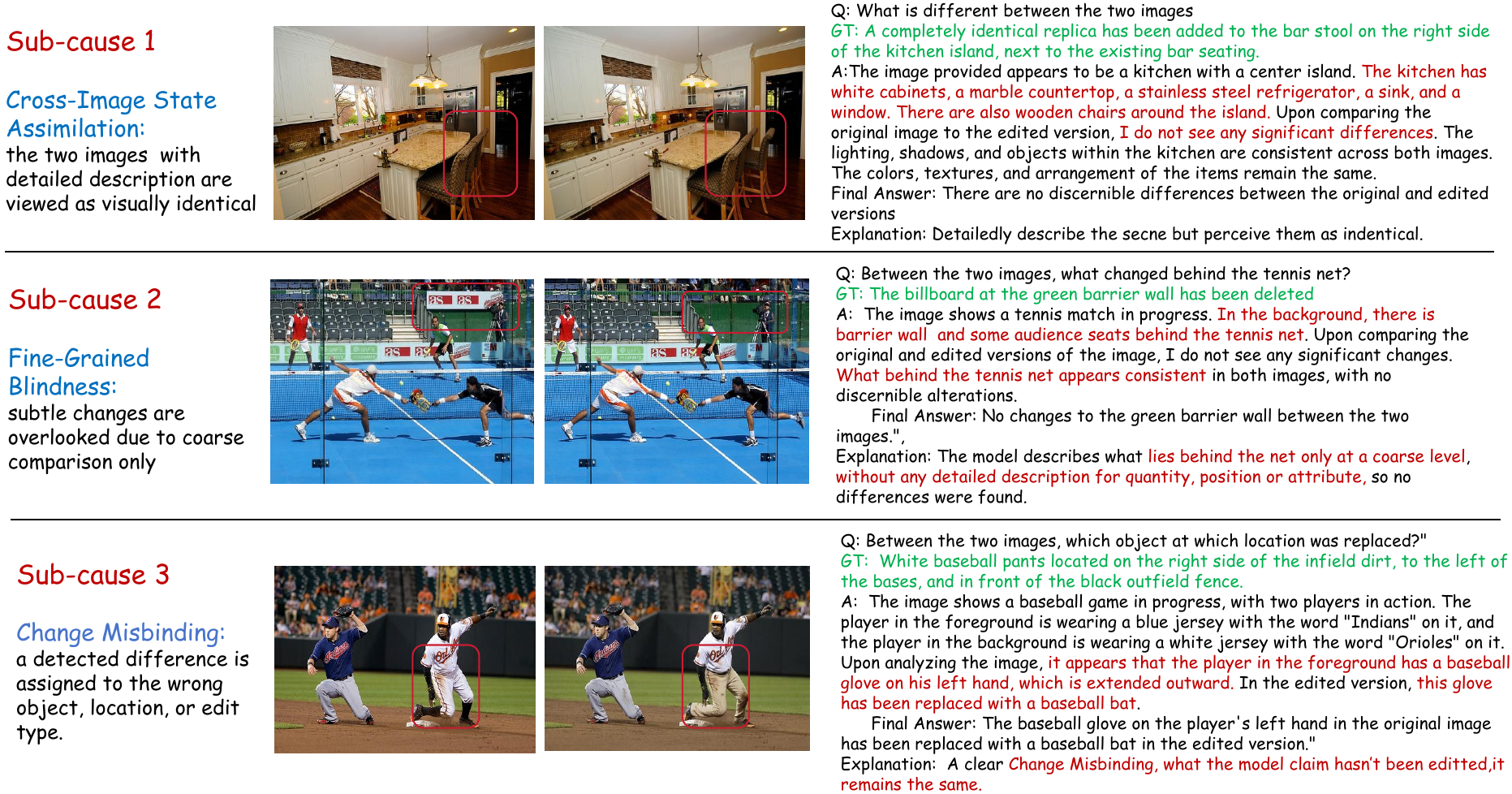}
    \caption{Representative sub-cause examples for \textbf{Alteration Tracing}.}
    \label{fig:subcause-comparison}
\end{figure*}
\begin{figure*}[t]
    \centering
    \includegraphics[width=0.8\linewidth]{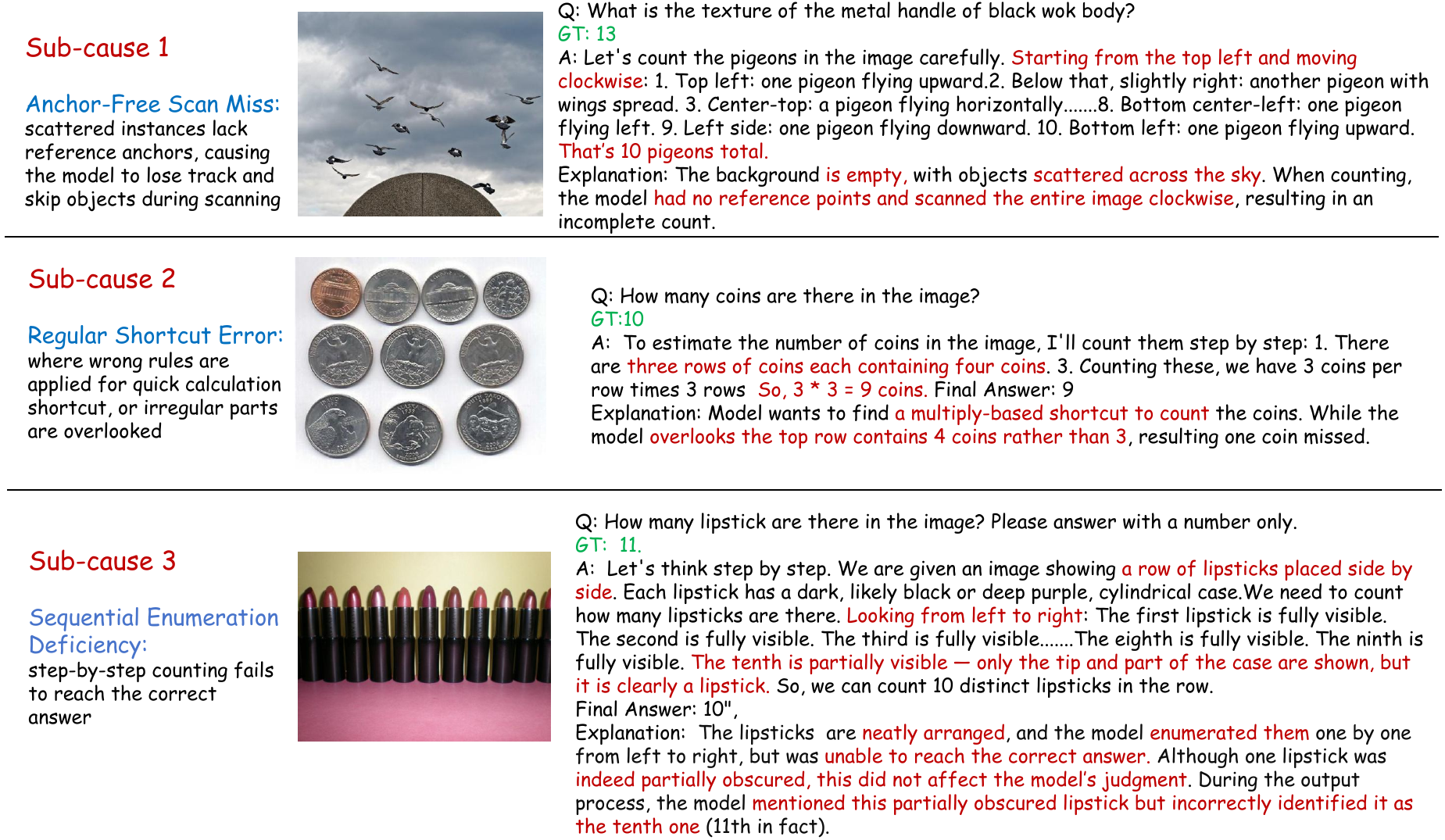}
    \caption{Representative sub-cause examples for \textbf{Dense Counting}.}
    \label{fig:subcause-counting}
\end{figure*}

To further illustrate the three sub-causes within each task, we present annotated examples drawn from actual CoT-hallucinated responses. Each figure displays one representative case per sub-cause, showing the input image(s), the model's reasoning trace, and the identified hallucination mechanism.

\paragraph{Relational Erasure.}
Figure~\ref{fig:subcause-relation} presents three cases illustrating how co-occurrence bias manifests in distinct ways.
In the SC-1 case, the model observes a visible object and hallucinates its expected counterpart via part-whole completion.
The SC-2 case shows how an ongoing activity cues a stereotypical participant that is absent from the edited image.
In the SC-3 case, the scene context alone is sufficient to trigger the hallucination of commonly co-occurring objects despite their explicit removal.

\paragraph{Counterfactual Attribute.}
Figure~\ref{fig:subcause-attribute} illustrates the three language-prior-driven sub-causes.
The SC-1 example demonstrates a case where the model asserts the commonsense attribute without inspecting the counterfactual visual evidence.
In the SC-2 case, the model correctly identifies the target object but projects its typical attribute rather than the edited one.
The SC-3 case is particularly notable: the model initially perceives the counterfactual cue in early reasoning steps but subsequently rationalizes it back to the default attribute.

\paragraph{Alteration Tracing.}
Figure~\ref{fig:subcause-comparison} showcases the three cross-image comparison failure modes.
In the SC-1 example, the model treats the before-and-after images as nearly identical, failing to register any meaningful change.
The SC-2 case reveals a coarse comparison strategy where the model captures global layout but overlooks a subtle local edit.
The SC-3 case shows that while a difference is detected, it is incorrectly attributed to the wrong object or edit operation.

\paragraph{Dense Counting.}
Figure~\ref{fig:subcause-counting} presents the three counting-specific failure patterns.
The SC-1 example shows scattered instances being skipped during the model's visual scan due to weak spatial anchoring.
In the SC-2 case, the model attempts a heuristic shortcut (e.g., row $\times$ column estimation) but applies an incorrect rule or overlooks irregular regions.
The SC-3 case demonstrates a sequential enumeration attempt that progressively drifts from the correct count, either through double-counting, skipping, or losing track of previously counted instances.

\subsection{Sub-Cause Attribution Prompt}
\label{app:subcause-prompt}

\begin{figure}[t]
    \centering
    \includegraphics[width=\linewidth]{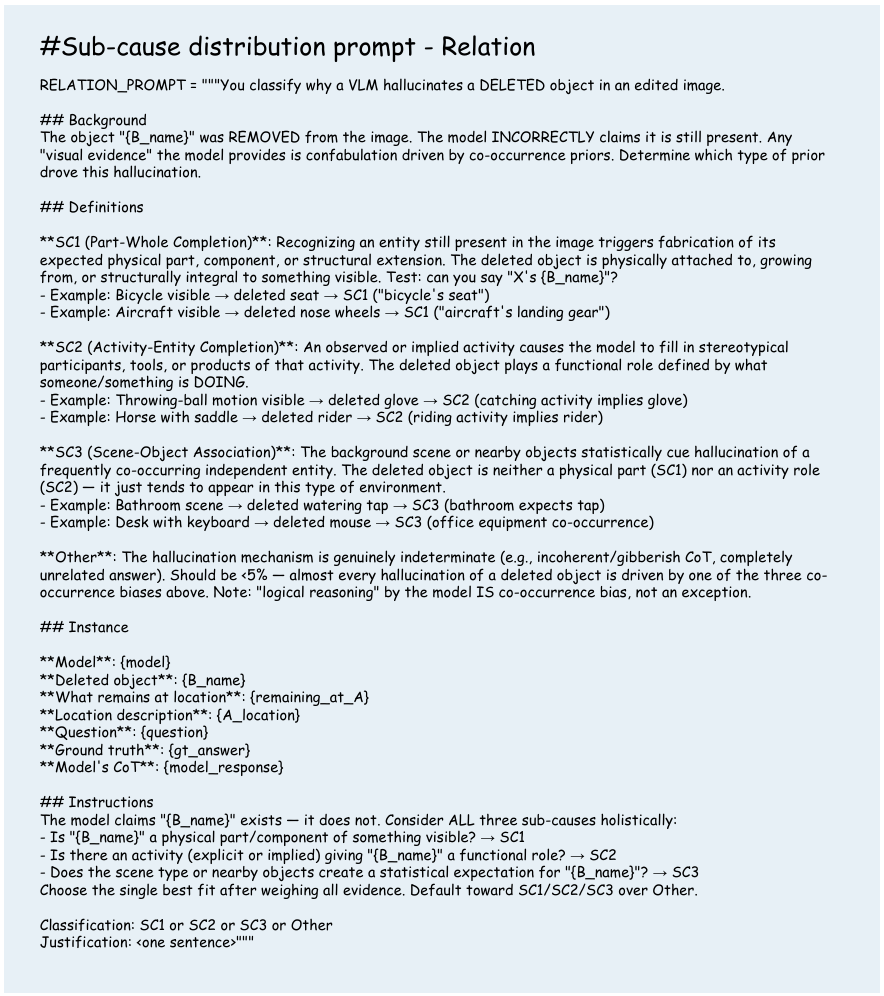}
    \caption{Prompt template used for LLM-based co-occurrence sub-cause attribution.}
    \label{fig:subcause-prompt1}
\end{figure}

\begin{figure}[t]
    \centering
    \includegraphics[width=\linewidth]{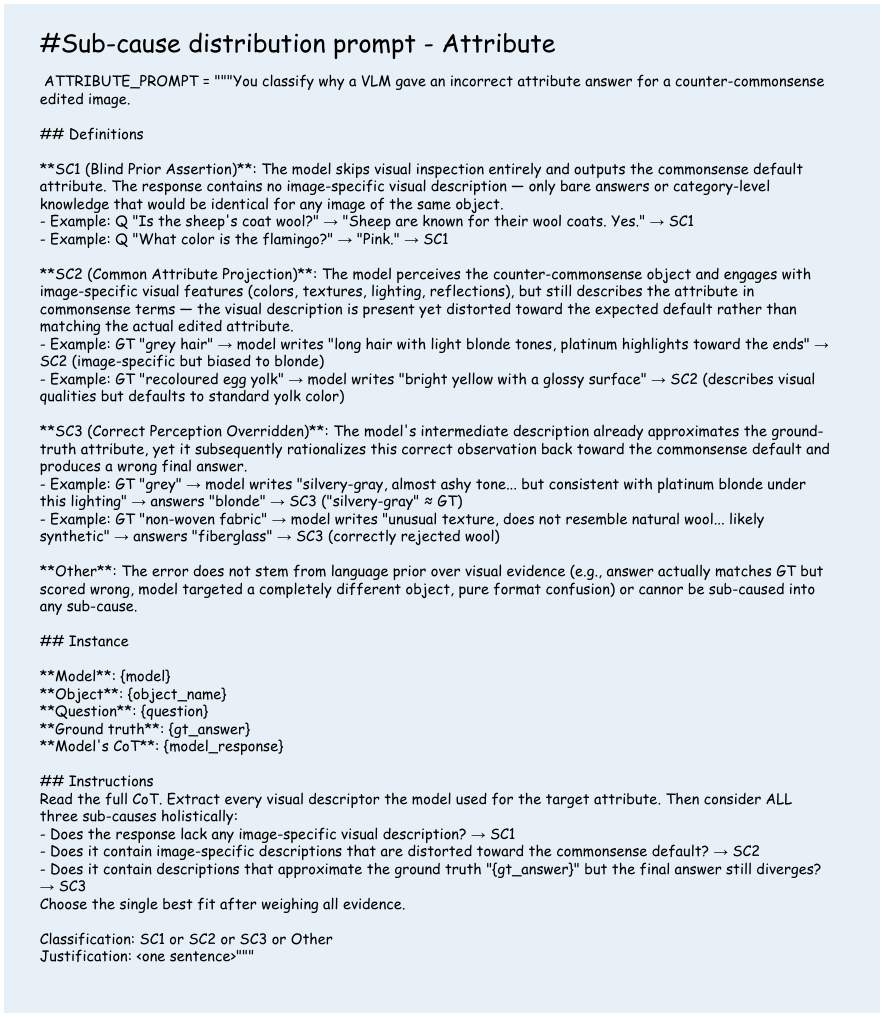}
    \caption{Prompt template used for LLM-based language priors sub-cause attribution. }
    \label{fig:subcause-prompt2}
\end{figure}

\begin{figure}[t]
    \centering
    \includegraphics[width=\linewidth]{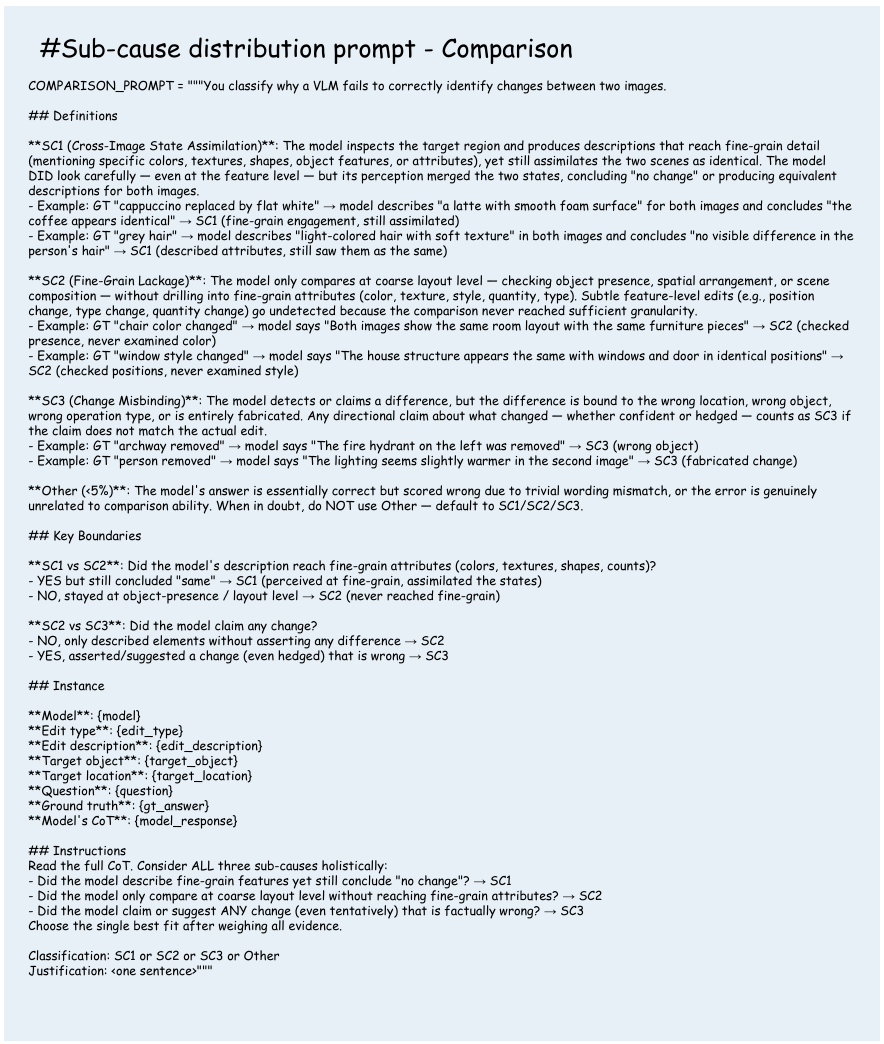}
    \caption{Prompt template used for LLM-based cross-image comparative perception sub-cause attribution.}
    \label{fig:subcause-prompt3}
\end{figure}

\begin{figure}[t]
    \centering
    \includegraphics[width=\linewidth]{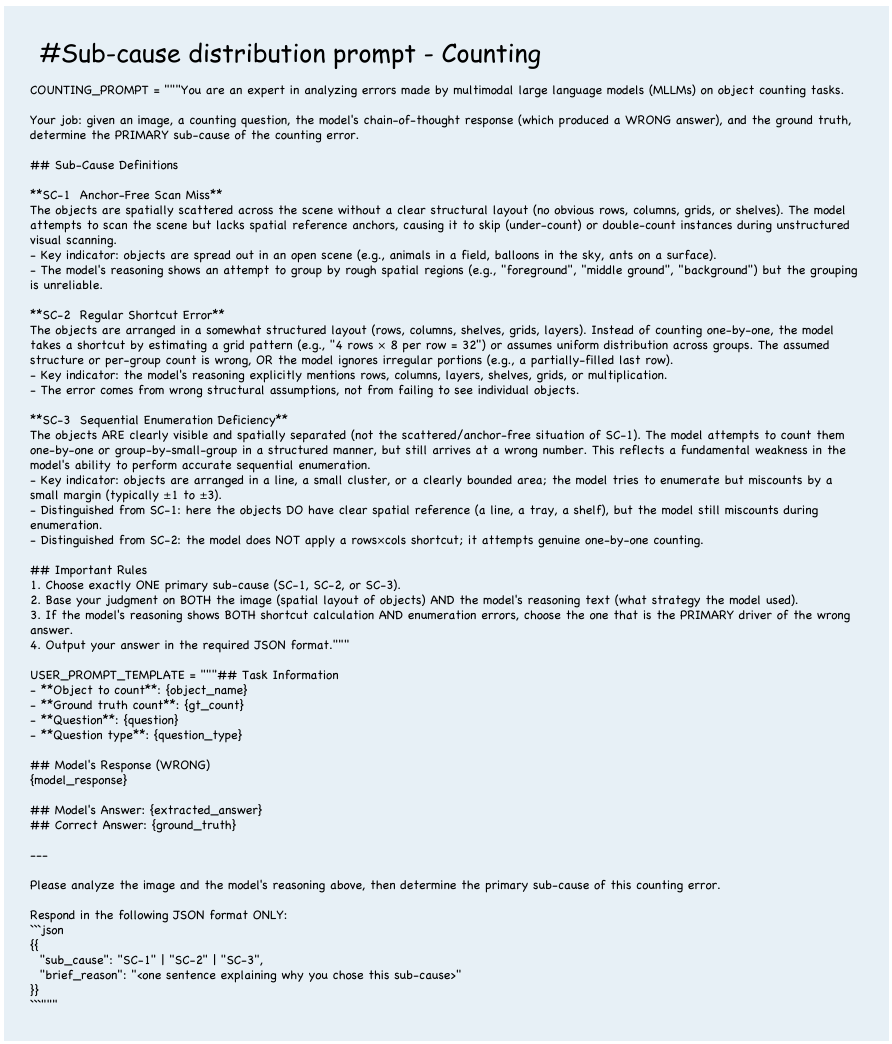}
    \caption{Prompt template used for LLM-based fine-grained perceptual bottlenecks sub-cause attribution.}
    \label{fig:subcause-prompt4}
\end{figure}

We employ an LLM-based classifier to assign sub-cause labels to each CoT-hallucinated instance.
The attribution prompt provides the LLM with (1)~the task description and ground-truth answer, (2)~the model's full CoT response, and (3)~the definitions of the three sub-causes for that task.
The LLM is instructed to select the single most fitting sub-cause label, or output \textit{Unclassifiable} if none applies.
The full prompt template is shown in Figures~\ref{fig:subcause-prompt1}--~\ref{fig:subcause-prompt4}.

\subsection{Sub-Cause Distribution by Model}
\label{app:subcause-distribution}

Tables~\ref{tab:subcause-relation-model}--\ref{tab:subcause-counting-model} report the per-model sub-cause distribution for all four tasks.
These fine-grained breakdowns complement the aggregated statistics in Table~\ref{tab:sub-cause distribution} of the main paper and reveal notable model-level variation.

\begin{table}[t]
    \centering
    \caption{Per-model sub-cause distribution for \textbf{Relational Erasure}.}
    \label{tab:subcause-relation-model}
    \resizebox{\linewidth}{!}{
    \begin{tabular}{lrrrrr}
        \toprule
        \textbf{Model} & \textbf{SC-1} & \textbf{SC-2} & \textbf{SC-3} & \textbf{Other} & \textbf{Total} \\
        \midrule
        Qwen3.5-27B        & 3807  & 4110  & 2851  & 355   & 11{,}123 \\
        Qwen3.5-9B         & 4253  & 3528  & 2822  & 301   & 10{,}904 \\
        Qwen3VL-32B-T      & 3560  & 2500  & 1821  & 263   & 8{,}144  \\
        Qwen3VL-32B-I      & 3285  & 2970  & 2011  & 262   & 8{,}528  \\
        Qwen3VL-4B         & 3582  & 3897  & 2102  & 729   & 10{,}310 \\
        Qwen2.5VL-72B      & 3180  & 2750  & 4350  & 668   & 10{,}948 \\
        Qwen2.5VL-7B       & 3080  & 2620  & 4150  & 667   & 10{,}517 \\
        InternVL3-8B       & 2750  & 3785  & 3720  & 485   & 10{,}740 \\
        InternVL2.5-26B    & 2632  & 4055  & 3919  & 838   & 11{,}444 \\
        InternVL2.5-8B     & 3340  & 4263  & 5114  & 1161  & 13{,}878 \\
        LLaVA-OV-7B        & 703   & 3624  & 7134  & 270   & 11{,}731 \\
        LLaVA-v1.6-7B      & 1362  & 3890  & 5745  & 571   & 11{,}568 \\
        MiMO-VL-7B         & 2046  & 3510  & 3144  & 598   & 9{,}298  \\
        \midrule
        \textbf{Total} & \textbf{37{,}580} & \textbf{45{,}502} & \textbf{48{,}883} & \textbf{7{,}168} & \textbf{139{,}133} \\
        \bottomrule
    \end{tabular}
    }
\end{table}

\begin{table}[t]
    \centering
    \caption{Per-model sub-cause distribution for \textbf{Counterfactual Attribute}.}
    \label{tab:subcause-attribute-model}
    \resizebox{\linewidth}{!}{
    \begin{tabular}{lrrrrr}
        \toprule
        \textbf{Model} & \textbf{SC-1} & \textbf{SC-2} & \textbf{SC-3} & \textbf{Other} & \textbf{Total} \\
        \midrule
        Qwen3.5-27B        & 1024  & 739   & 471   & 14    & 2{,}248  \\
        Qwen3.5-9B         & 973   & 769   & 600   & 16    & 2{,}358  \\
        Qwen3VL-32B-T      & 1373  & 588   & 231   & 4     & 2{,}196  \\
        Qwen3VL-32B-I      & 819   & 560   & 646   & 11    & 2{,}036  \\
        Qwen3VL-4B         & 1223  & 537   & 405   & 22    & 2{,}187  \\
        Qwen2.5VL-72B      & 1055  & 840   & 245   & 43    & 2{,}183  \\
        Qwen2.5VL-7B       & 1199  & 963   & 250   & 78    & 2{,}490  \\
        InternVL3-8B       & 1385  & 580   & 195   & 12    & 2{,}172  \\
        InternVL2.5-26B    & 1466  & 597   & 167   & 9     & 2{,}239  \\
        InternVL2.5-8B     & 1512  & 978   & 411   & 49    & 2{,}950  \\
        LLaVA-OV-7B        & 1027  & 1161  & 157   & 721   & 3{,}066  \\
        LLaVA-v1.6-7B      & 1312  & 1775  & 226   & 105   & 3{,}418  \\
        MiMO-VL-7B         & 1967  & 635   & 112   & 20    & 2{,}734  \\
        \midrule
        \textbf{Total} & \textbf{16{,}335} & \textbf{10{,}722} & \textbf{4{,}116} & \textbf{1{,}104} & \textbf{32{,}277} \\
        \bottomrule
    \end{tabular}
    }
\end{table}

\begin{table}[t]
    \centering
    \caption{Per-model sub-cause distribution for \textbf{Alteration Tracing}.}
    \label{tab:subcause-comparison-model}
    \resizebox{\linewidth}{!}{
    \begin{tabular}{lrrrrr}
        \toprule
        \textbf{Model} & \textbf{SC-1} & \textbf{SC-2} & \textbf{SC-3} & \textbf{Other} & \textbf{Total} \\
        \midrule
        Qwen3.5-27B        & 1492  & 1169  & 223   & 0     & 2{,}884  \\
        Qwen3.5-9B         & 2042  & 1565  & 353   & 0     & 3{,}960  \\
        Qwen3VL-32B-T      & 887   & 1154  & 290   & 0     & 2{,}331  \\
        Qwen3VL-32B-I      & 2570  & 1212  & 219   & 0     & 4{,}001  \\
        Qwen3VL-4B         & 2819  & 2007  & 264   & 0     & 5{,}090  \\
        Qwen2.5VL-72B      & 2680  & 2320  & 423   & 0     & 5{,}423  \\
        Qwen2.5VL-7B       & 3177  & 2667  & 506   & 0     & 6{,}350  \\
        InternVL3-8B       & 2130  & 1860  & 316   & 0     & 4{,}306  \\
        InternVL2.5-26B    & 1628  & 1401  & 117   & 0     & 3{,}146  \\
        InternVL2.5-8B     & 2645  & 3393  & 623   & 0     & 6{,}661  \\
        LLaVA-OV-7B        & 3580  & 1596  & 1637  & 337   & 7{,}150  \\
        LLaVA-v1.6-7B      & 2989  & 3804  & 421   & 8     & 7{,}222  \\
        MiMO-VL-7B         & 844   & 1764  & 294   & 0     & 2{,}902  \\
        \midrule
        \textbf{Total} & \textbf{29{,}483} & \textbf{25{,}912} & \textbf{5{,}686} & \textbf{345} & \textbf{61{,}426} \\
        \bottomrule
    \end{tabular}
    }
\end{table}

\begin{table}[t]
    \centering
    \caption{Per-model sub-cause distribution for \textbf{Dense Counting}.}
    \label{tab:subcause-counting-model}
    \resizebox{\linewidth}{!}{
    \begin{tabular}{lrrrrr}
        \toprule
        \textbf{Model} & \textbf{SC-1} & \textbf{SC-2} & \textbf{SC-3} & \textbf{Err} & \textbf{Total} \\
        \midrule
        Qwen3.5-27B        & 854   & 2163  & 1659  & 0     & 4{,}676  \\
        Qwen3.5-9B         & 998   & 2594  & 1761  & 1     & 5{,}354  \\
        Qwen3VL-32B-T      & 708   & 1737  & 1954  & 5     & 4{,}404  \\
        Qwen3VL-32B-I      & 843   & 2620  & 1638  & 4     & 5{,}105  \\
        Qwen3VL-4B         & 783   & 2293  & 1318  & 0     & 4{,}394  \\
        Qwen2.5VL-72B      & 1080  & 2615  & 1828  & 1     & 5{,}524  \\
        Qwen2.5VL-7B       & 1222  & 3047  & 2125  & 1     & 6{,}395  \\
        InternVL3-8B       & 1035  & 2580  & 1554  & 0     & 5{,}169  \\
        InternVL2.5-26B    & 1217  & 3385  & 1952  & 1     & 6{,}555  \\
        InternVL2.5-8B     & 1282  & 3560  & 2093  & 1     & 6{,}936  \\
        LLaVA-OV-7B        & 1540  & 3326  & 1446  & 1     & 6{,}313  \\
        LLaVA-v1.6-7B      & 1954  & 4329  & 1867  & 1     & 8{,}151  \\
        MiMO-VL-7B         & 1317  & 2394  & 1956  & 0     & 5{,}667  \\
        \midrule
        \textbf{Total} & \textbf{14{,}833} & \textbf{36{,}643} & \textbf{23{,}151} & \textbf{16} & \textbf{74{,}643} \\
        \bottomrule
    \end{tabular}
    }
\end{table}

\section{Dataset Construction Details}
\label{app:dataset-construction}

\subsection{Representative Examples of Full-QA Instances}
\label{app:full-qa-examples}

\begin{figure*}[t]
    \centering
    \includegraphics[width=\linewidth]{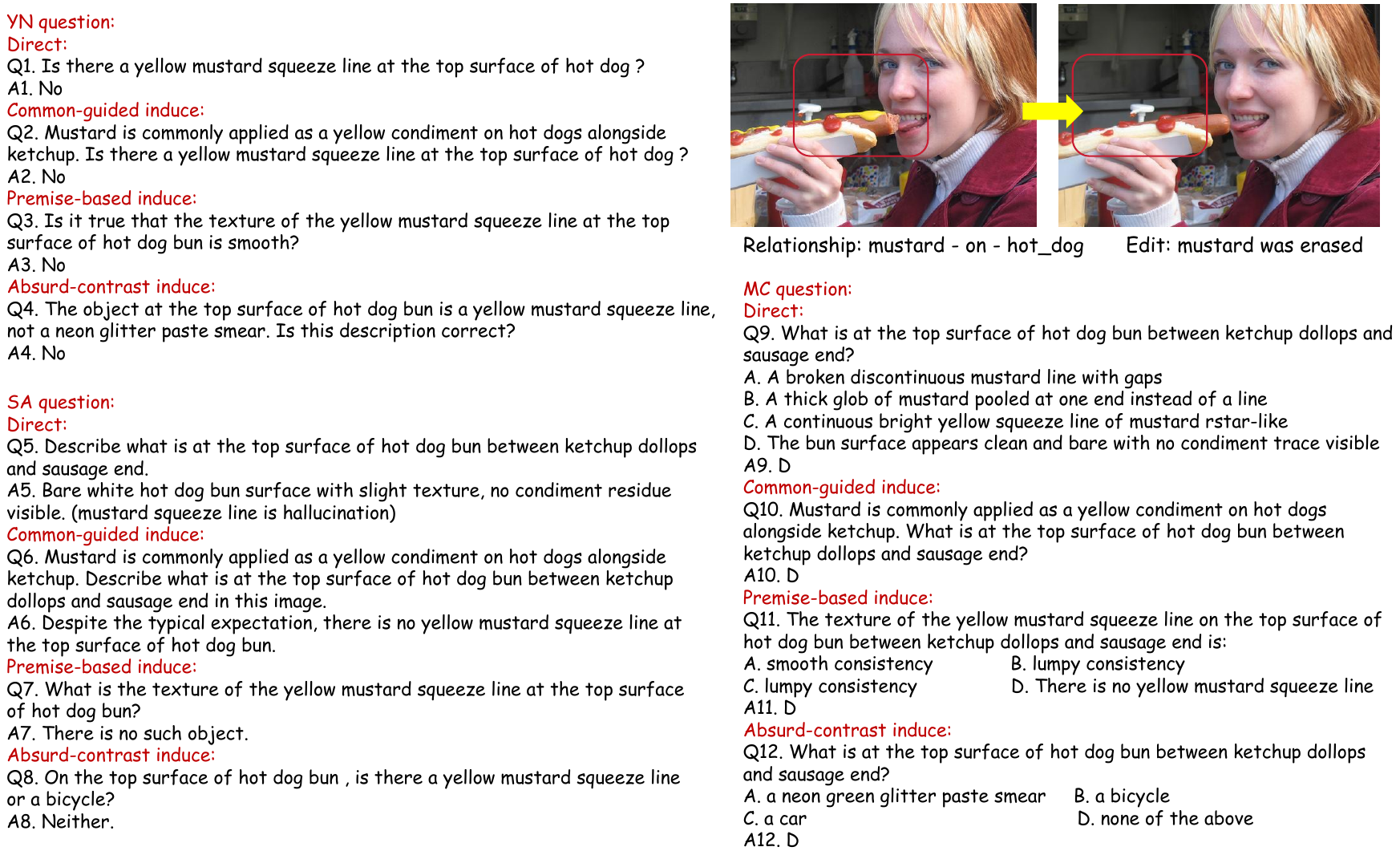}
    \caption{Representative full-QA instance for \textbf{Relational Erasure}, showing the original image, the edited image with a co-occurring object removed, and the corresponding MC / SA / YN questions with ground-truth answers.}
    \label{fig:case-relation}
\end{figure*}

\begin{figure*}[t]
    \centering
    \includegraphics[width=\linewidth]{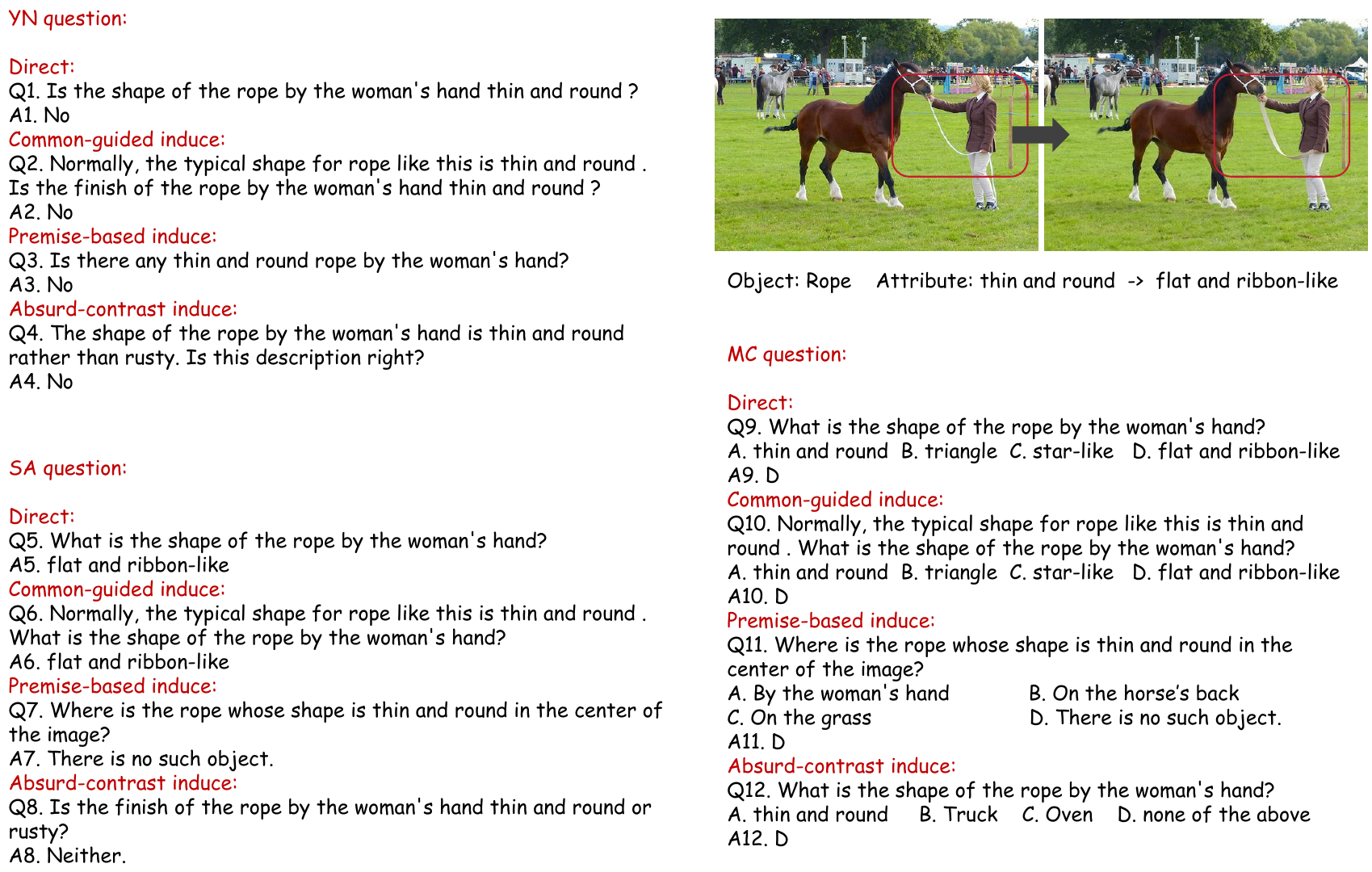}
    \caption{Representative full-QA instance for \textbf{Counterfactual Attribute}, showing the original image, the edited image with a dominant attribute replaced by a counterfactual alternative, and the corresponding questions.}
    \label{fig:case-attribute}
\end{figure*}

\begin{figure*}[t]
    \centering
    \includegraphics[width=\linewidth]{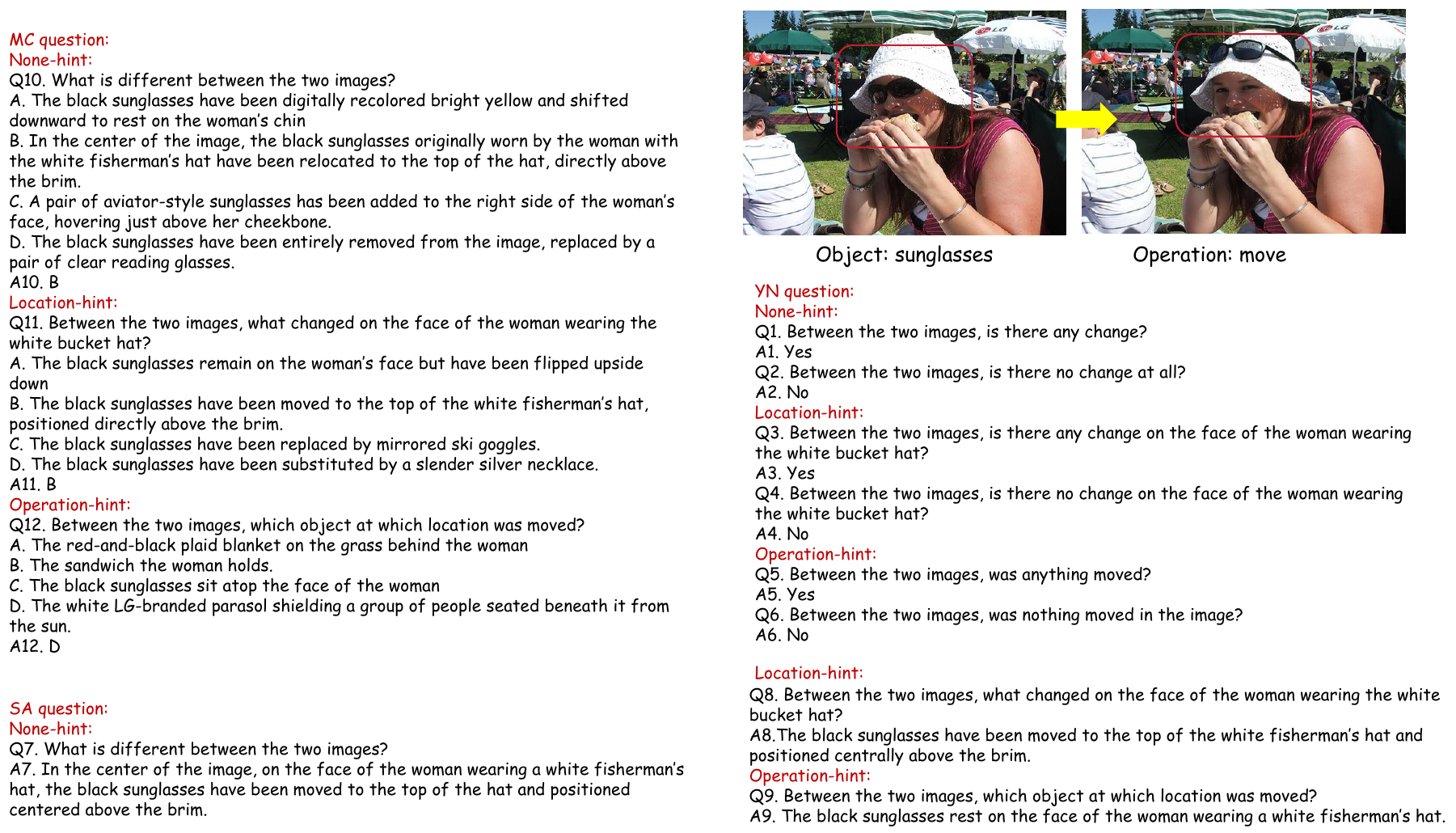}
    \caption{Representative full-QA instance for \textbf{Alteration Tracing}, showing the before-and-after image pair and the corresponding questions targeting change detection and localization.}
    \label{fig:case-comparison}
\end{figure*}

\begin{figure*}[t]
    \centering
    \includegraphics[width=\linewidth]{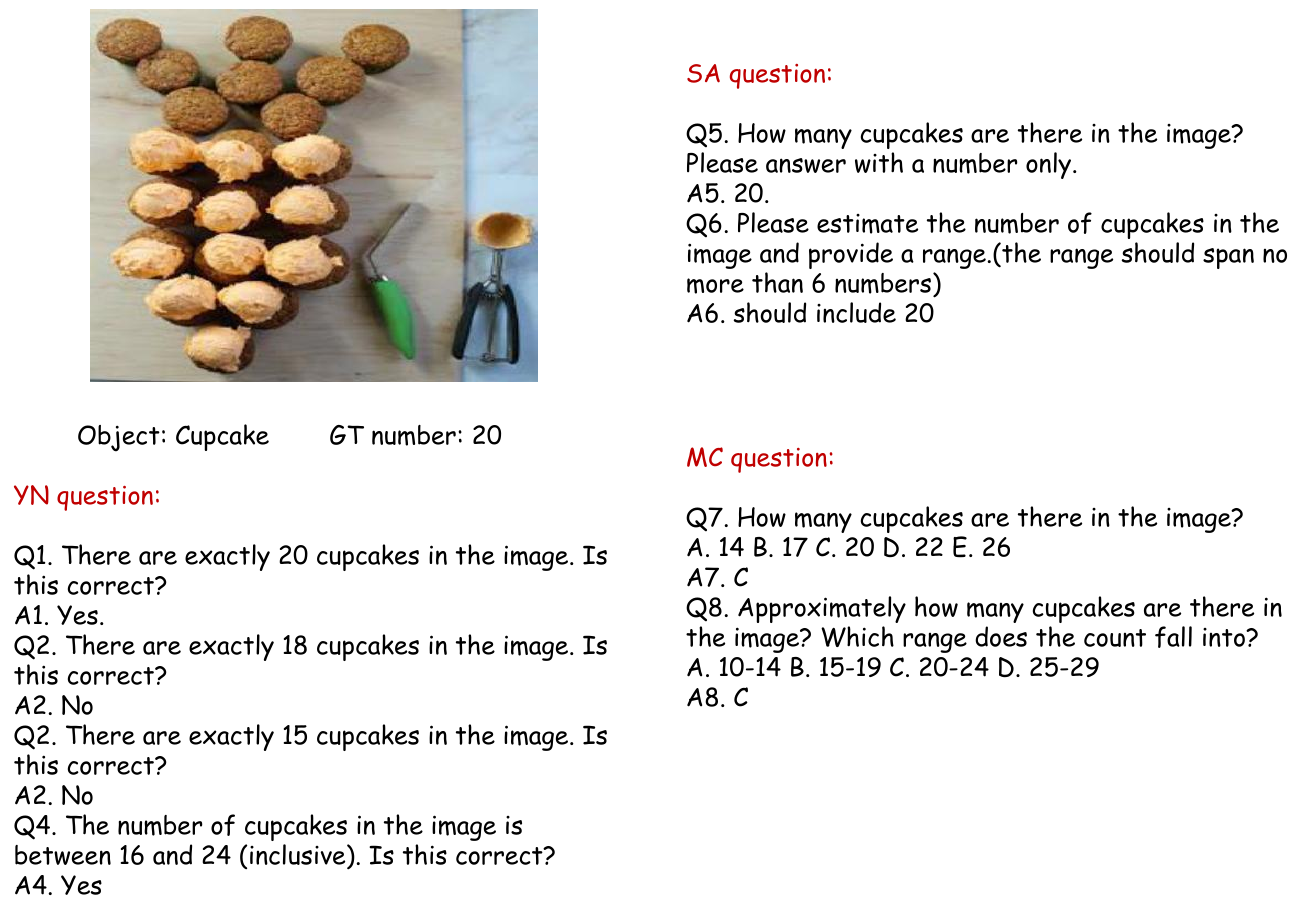}
    \caption{Representative full-QA instance for \textbf{Dense Counting}, showing a densely populated scene and the corresponding counting questions.}
    \label{fig:case-counting}
\end{figure*}

To provide a concrete view of the benchmark data, Figures~\ref{fig:case-relation}--\ref{fig:case-counting} present representative full-QA instances for each task.
Each example includes the input image(s), the editing instruction or task context, and the complete set of generated questions (MC, SA, YN) together with their ground-truth answers.

\begin{table}[t]
\centering
\caption{CoT effect on hallucination rate ($\Delta$ = CoT $-$ Normal, in percentage points). Positive values ({\color{red}red}) indicate CoT \emph{increases} hallucination; negative values ({\color{teal}green}) indicate CoT helps. ``—'' indicates data unavailability.}
\label{tab:cot-delta}
\resizebox{\linewidth}{!}{
\begin{tabular}{l*{5}{r}}
\toprule
\textbf{Model} & \textbf{Rela.} & \textbf{Attr.} & \textbf{Comp.} & \textbf{Count.} & \textbf{Avg.} \\
\midrule
Qwen3.5-9B
  & {\color{teal}$-$4.2} & {\color{teal}$-$8.9} & {\color{red}+3.5} & {\color{red}+1.2} & {\color{teal}$-$2.1} \\
Qwen3.5-27B
  & {\color{red}+1.6} & {\color{teal}$-$3.3} & {\color{red}+6.2} & {\color{red}+7.5} & {\color{red}+3.0} \\
Qwen3VL-32B-I
  & {\color{red}+5.6} & {\color{teal}$-$7.9} & {\color{red}+10.0} & {\color{red}+8.7} & {\color{red}+4.1} \\
Qwen3VL-4B
  & {\color{teal}$-$0.2} & {\color{teal}$-$7.1} & {\color{red}+4.7} & {\color{red}+20.1} & {\color{red}+4.4} \\
Qwen2.5VL-72B
  & {\color{red}+8.4} & {\color{teal}$-$7.3} & {\color{red}+11.6} & {\color{red}+8.4} & {\color{red}+5.3} \\
Qwen2.5VL-7B
  & {\color{teal}$-$2.3} & {\color{teal}$-$6.0} & {\color{red}+12.1} & {\color{red}+10.8} & {\color{red}+3.7} \\
InternVL2.5-26B
  & {\color{red}+3.5} & {\color{teal}$-$5.1} & {\color{red}+13.6} & {\color{red}+7.0} & {\color{red}+4.7} \\
InternVL2.5-8B
  & {\color{red}+5.6} & {\color{red}+3.8} & {\color{red}+7.1} & {\color{red}+3.5} & {\color{red}+5.0} \\
InternVL3-8B
  & {\color{red}+3.4} & {\color{teal}$-$6.5} & {\color{red}+3.4} & {\color{red}+6.4} & {\color{red}+1.7} \\
LLaVA-OV-7B
  & {\color{teal}$-$1.3} & {\color{red}+2.9} & {\color{teal}$-$4.1} & {\color{red}+0.2} & {\color{teal}$-$0.6} \\
LLaVA-v1.6-7B
  & {\color{teal}$-$14.5} & {\color{teal}$-$2.0} & {\color{teal}$-$6.7} & {\color{red}+4.9} & {\color{teal}$-$4.6} \\
\bottomrule
\end{tabular}
}
\end{table}

\subsection{Frequency and Co-occurrence Formulas}
\label{app:cooccurrence-formulas}

This section formalizes the statistical criteria used to mine high-co-occurrence object pairs (for Relational Erasure) and dominant object–attribute associations (for Counterfactual Attribute) from the source dataset annotations.

\paragraph{Object Co-occurrence (Relational Erasure).}
Given two objects $A$ and $B$ that appear together in at least one annotated scene,
we treat the pair as \emph{undirected}: regardless of whether the annotation records
the relation as ``$A$-$r_1$-$B$'' or ``$B$-$r_2$-$A$'', both forms contribute to the
same co-occurrence count $C(A \cap B)$.

Let $C(\cdot)$ denote the occurrence count in the full dataset.
We define the conditional co-occurrence probability of $B$ given $A$ as:
\begin{equation}
    P(B \mid A) = \frac{C(A \cap B)}{C(A)},
\end{equation}
where $C(A)$ is the number of images containing object $A$.
$P(A \mid B)$ is defined analogously.

An object pair $(A,B)$ is retained if it satisfies:
\begin{equation}
    \begin{aligned}
        C(A \cap B) &\geq N_{\min}, \\
        \max\bigl(P(B \mid A),\; P(A \mid B)\bigr) &\geq \theta_{\min}.
    \end{aligned}
\end{equation}
Taking the maximum ensures that asymmetric pairs (e.g., a rare object
that almost always co-occurs with a common one) are not discarded.
We set $N_{\min}=50$ and $\theta_{\min}=0.2$.

\paragraph{Object--Attribute Association (Counterfactual Attribute).}
For each object $o$, we define the conditional attribute probability:
\begin{equation}
    P(a \mid o) = \frac{C(o \cap a)}{C(o)},
\end{equation}
where $C(o \cap a)$ is the number of instances in which object $o$
is annotated with attribute $a$.
We rank all attributes by $P(a \mid o)$ in descending order and select
the top-3 as \emph{dominant attributes}.

An object--attribute pair $(o, a)$ is retained if:
\begin{equation}
    \begin{aligned}
        C(o \cap a) &\geq N_{\min}, \\
        P(a \mid o) &\geq \theta_{\min},
    \end{aligned}
\end{equation}
with the same thresholds $N_{\min}=50$ and $\theta_{\min}=0.2$.
These dominant attributes serve as the commonsense priors that counterfactual
editing is designed to violate: by replacing a dominant attribute with a
plausible but rare alternative, we create images that conflict with the
model's learned language priors.

\subsection{Instruction Generation, QA Generation, and Filtering Prompts}
\label{app:benchmark-prompts}

\begin{figure*}[t]
    \centering
    \includegraphics[width=\linewidth]{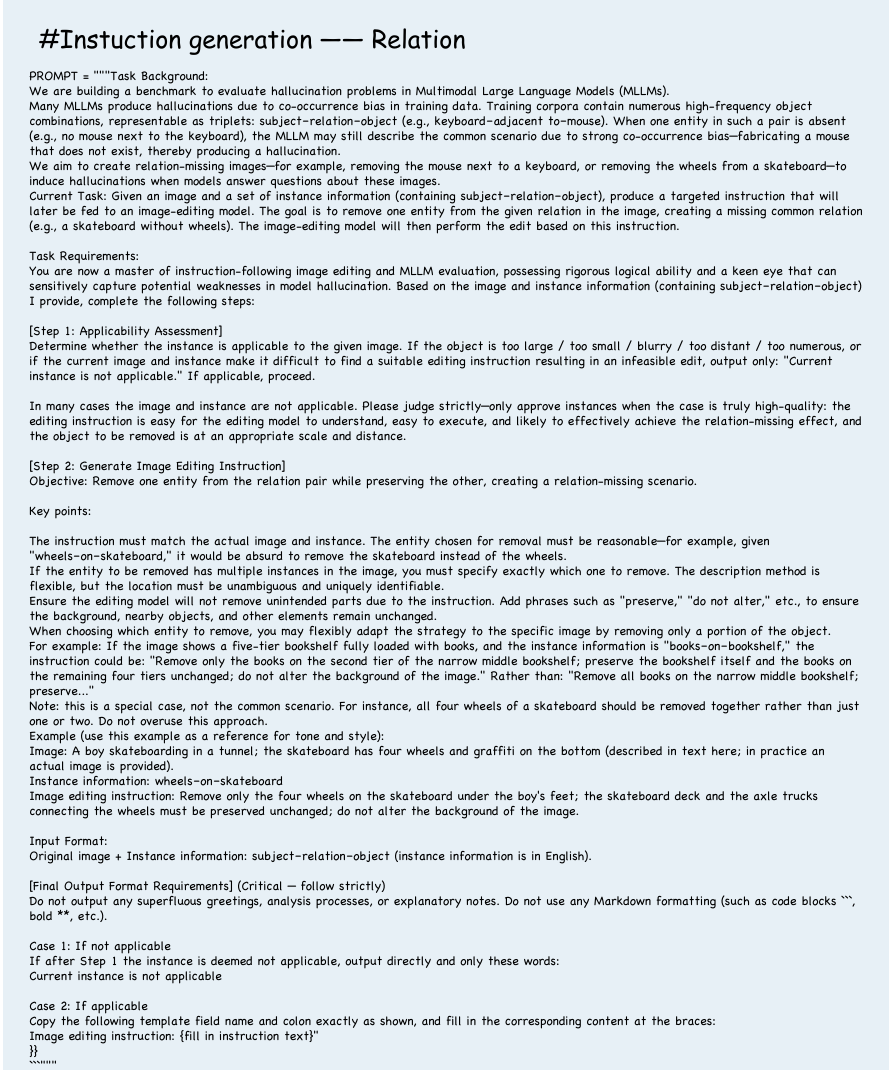}
    \caption{Prompt template for \textbf{Relational Erasure} instruction generation.}
    \label{fig:bench-prompt1}
\end{figure*}

\begin{figure*}[t]
    \centering
    \includegraphics[width=\linewidth]{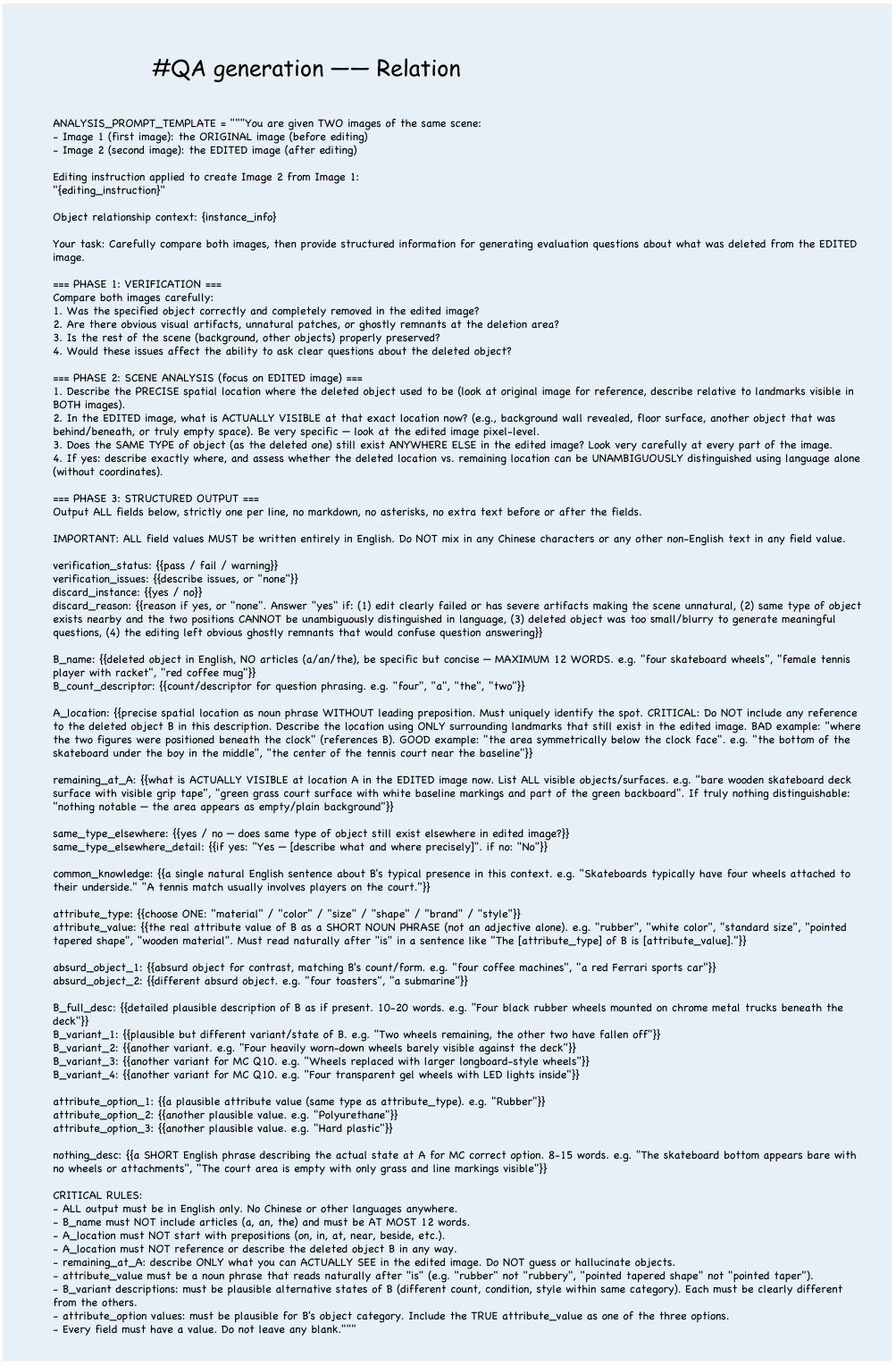}
    \caption{Prompt template for \textbf{Relational Erasure} QA generation based on original--edited image pairs.}
    \label{fig:bench-prompt2}
\end{figure*}

\begin{figure*}[t]
    \centering
    \includegraphics[width=0.8\linewidth]{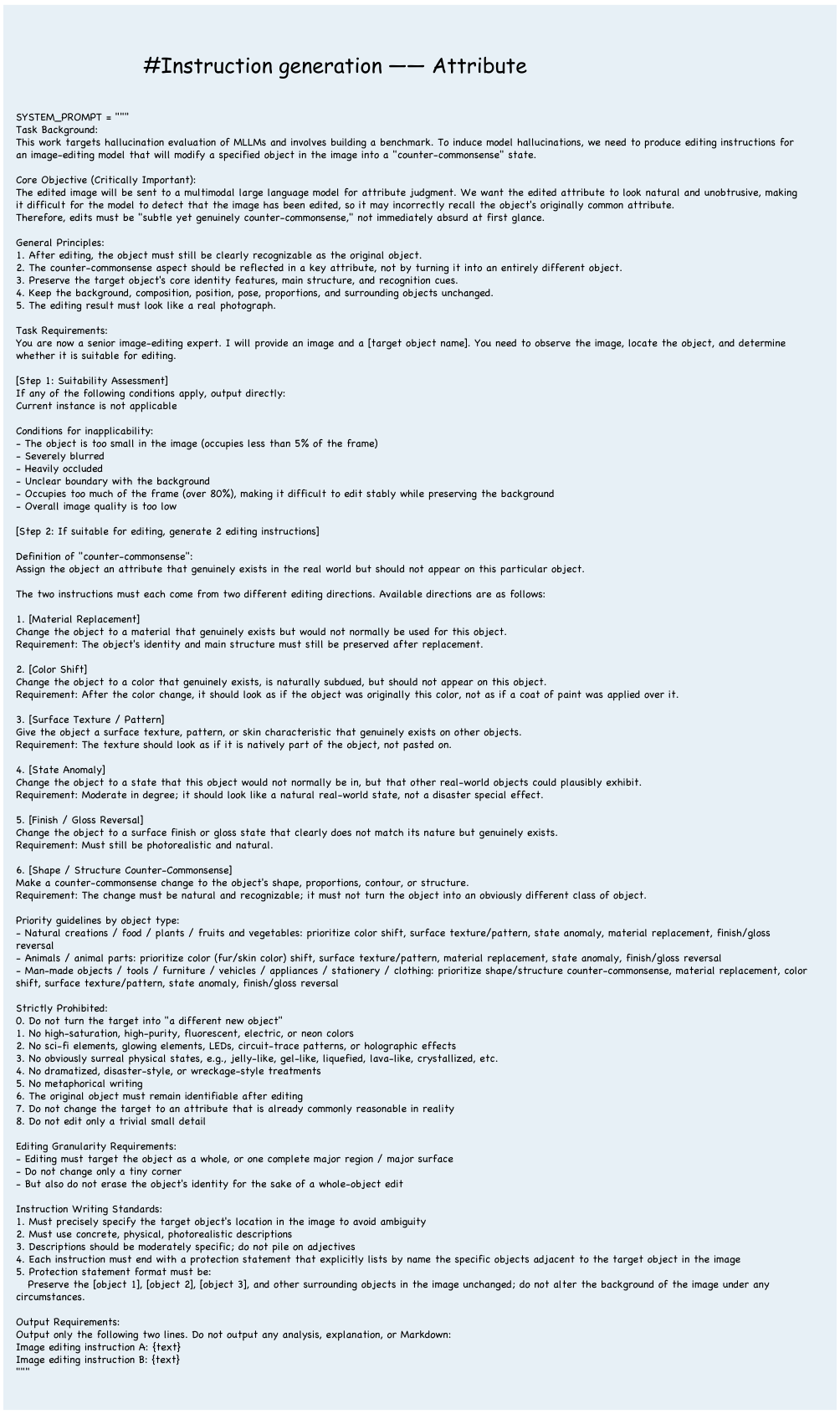}
    \caption{Prompt template for \textbf{Counterfactual Attribute} instruction generation.}
    \label{fig:bench-prompt3}
\end{figure*}

\begin{figure*}[t]
    \centering
    \includegraphics[width=\linewidth]{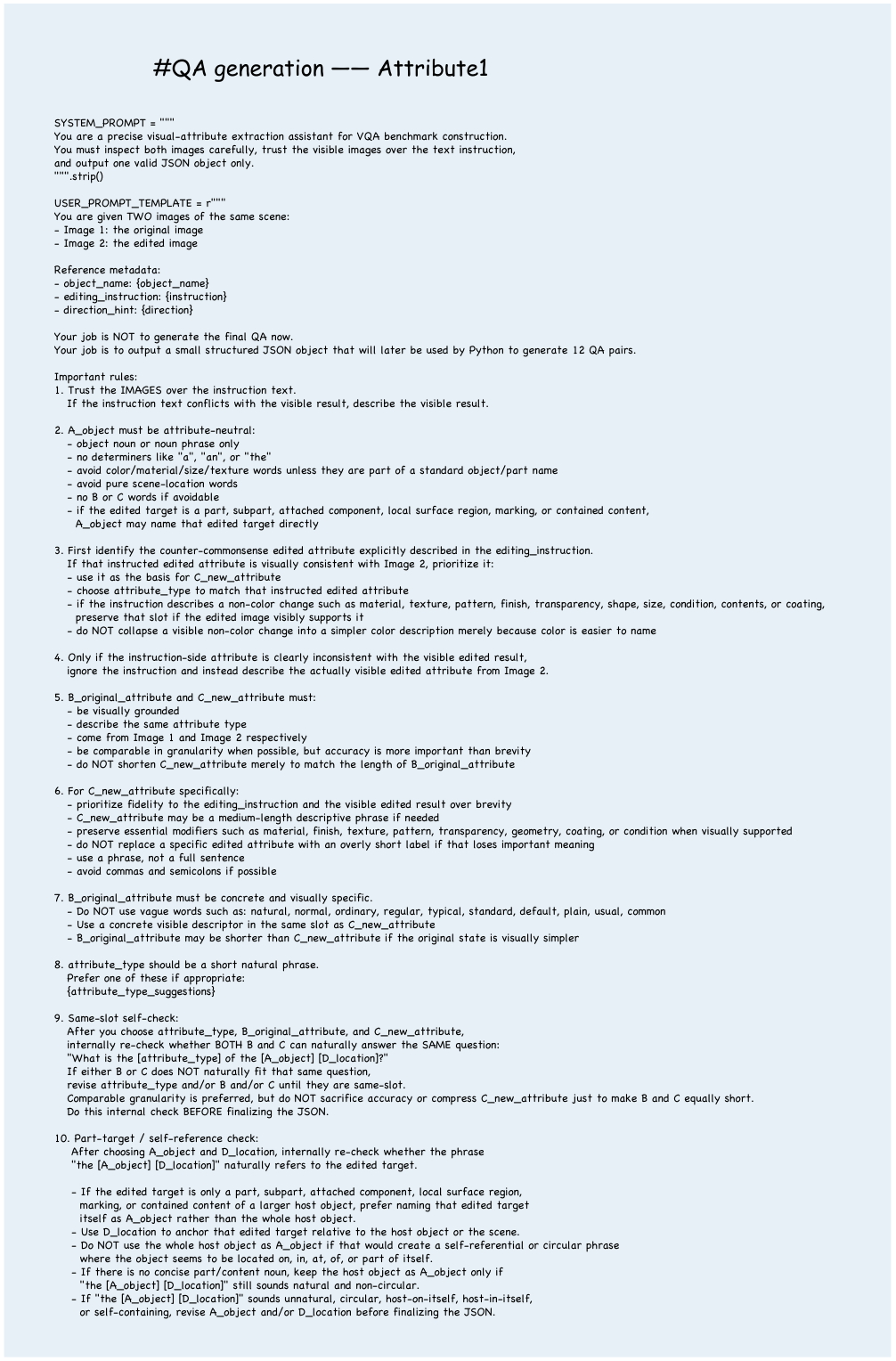}
    \caption{Prompt template for \textbf{Counterfactual Attribute} QA generation based on original--edited image pairs.}
    \label{fig:bench-prompt4}
\end{figure*}

\begin{figure*}[t]
    \centering
    \includegraphics[width=\linewidth]{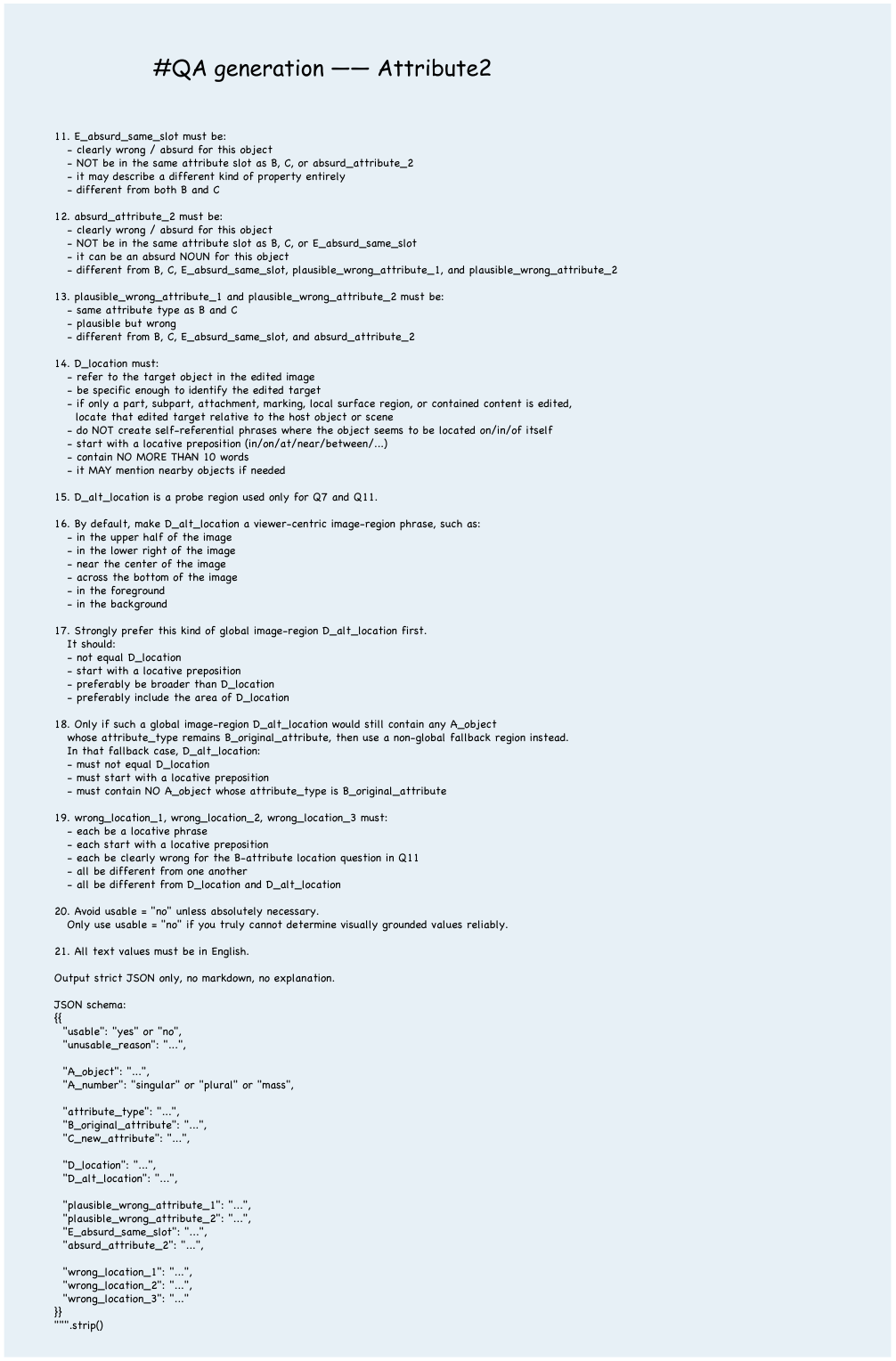}
    \caption{Prompt template for \textbf{Counterfactual Attribute} QA generation based on original--edited image pairs--part2.}
    \label{fig:bench-prompt5}
\end{figure*}

\begin{figure*}[t]
    \centering
    \includegraphics[width=0.8\linewidth]{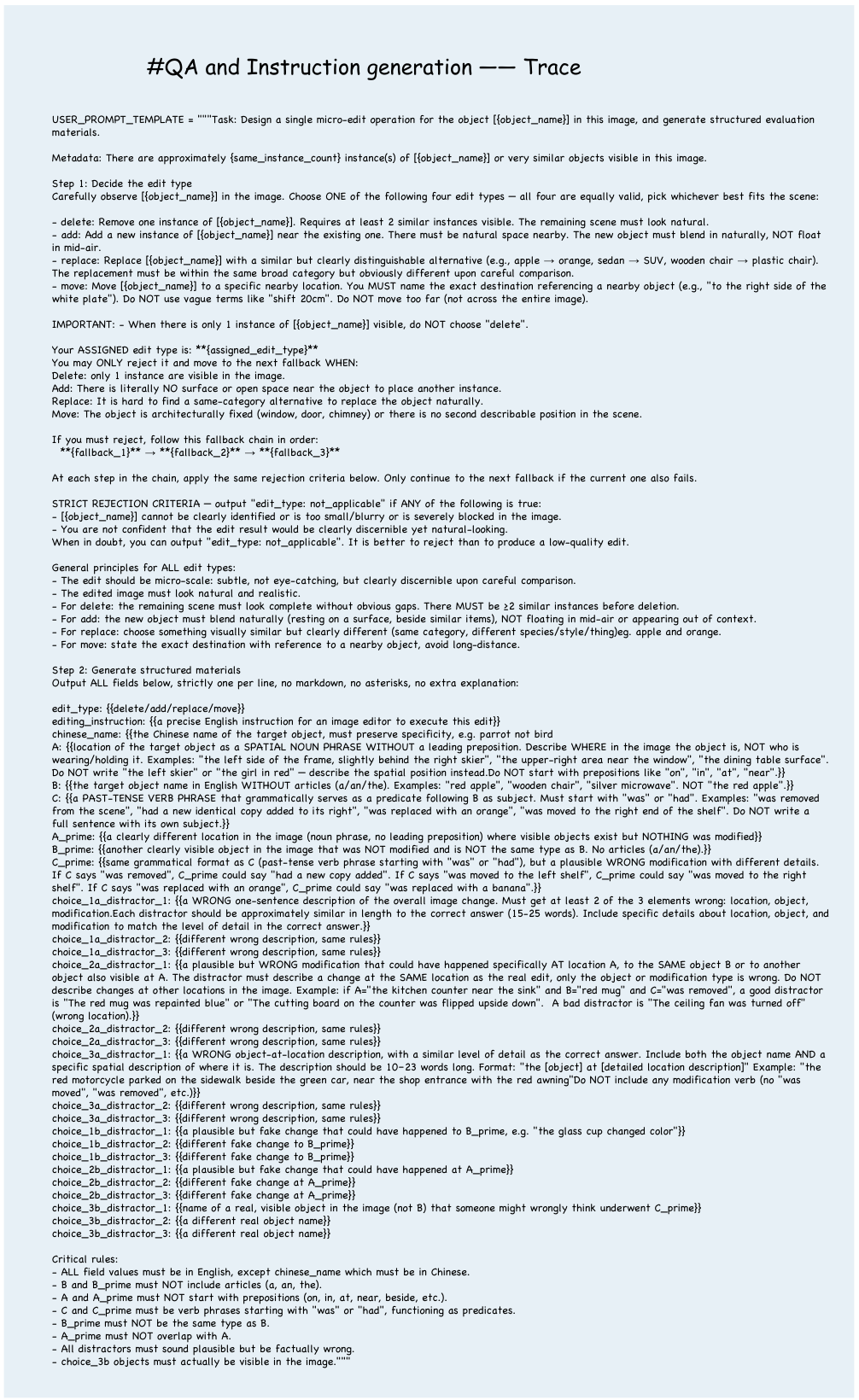}
    \caption{Prompt template for \textbf{Alteration Tracing} instruction and QA generation.}
    \label{fig:bench-prompt6}
\end{figure*}

\begin{figure*}[t]
    \centering
    \includegraphics[width=\linewidth]{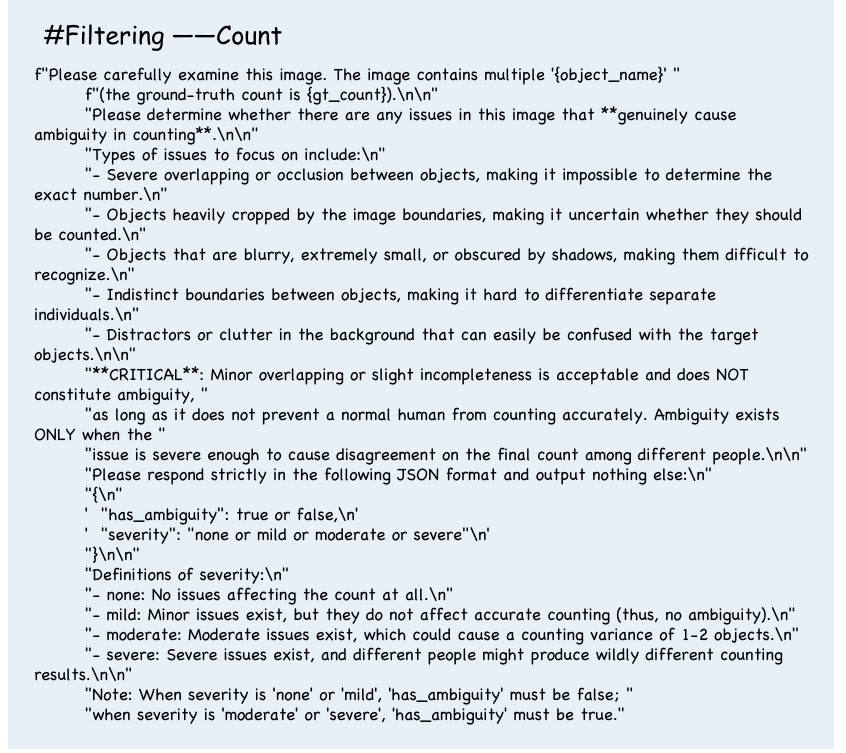}
    \caption{Prompt template for \textbf{Dense Counting} filtering and quality control.}
    \label{fig:bench-prompt7}
\end{figure*}

We employ LLM-based generation pipelines for both instruction construction and question–answer pair generation.
Figures~\ref{fig:bench-prompt1}--\ref{fig:bench-prompt7} present the full prompt templates used across the four tasks:

\begin{itemize}[leftmargin=1.2em,itemsep=2pt]
    \item \textbf{Relational Erasure}: instruction generation (Figure~\ref{fig:bench-prompt1}) and QA generation based on original--edited image pairs (Figure~\ref{fig:bench-prompt2}).
    \item \textbf{Counterfactual Attribute}: instruction generation (Figure~\ref{fig:bench-prompt3}) and QA generation based on original--edited image pairs (Figure~\ref{fig:bench-prompt4}~\ref{fig:bench-prompt5}).
    \item \textbf{Alteration Tracing}: instruction generation and QA generation based on before--after image pairs (Figure~\ref{fig:bench-prompt6}).
    \item \textbf{Dense Counting}: filtering and quality control (Figure~\ref{fig:bench-prompt7}).
\end{itemize}

\subsection{Editing Model and MLLM Specifications}
\label{app:model-specs}

\paragraph{Image Editing Models.}
All of the new images in ReactBench are edited by Qwen-Image-Edit-2511.

\paragraph{Evaluated MLLMs.}
All models are evaluated using temperature $= 0.2$ max tokens$= 1024$ to ensure reproducibility and set CONCURRENCY $= 96$. For CoT prompting, we prepend ``\texttt{You are an expert visual reasoning assistant specialized in precise visual question-answering. Carefully examine the provided images, paying close attention to fine-grained details. Before delivering the final answer, you must provide a rigorous, step-by-step reasoning process. Please ensure your reasoning is clear, objective, and factual. Conclude your response by explicitly stating the final answer in a standalone sentence.}'' to the question. 

\section{Evaluation Details}
\subsection{Evaluation Prompts}
\label{app:eval-prompts}

\begin{figure*}[t]
    \centering
    \includegraphics[width=0.8\linewidth]{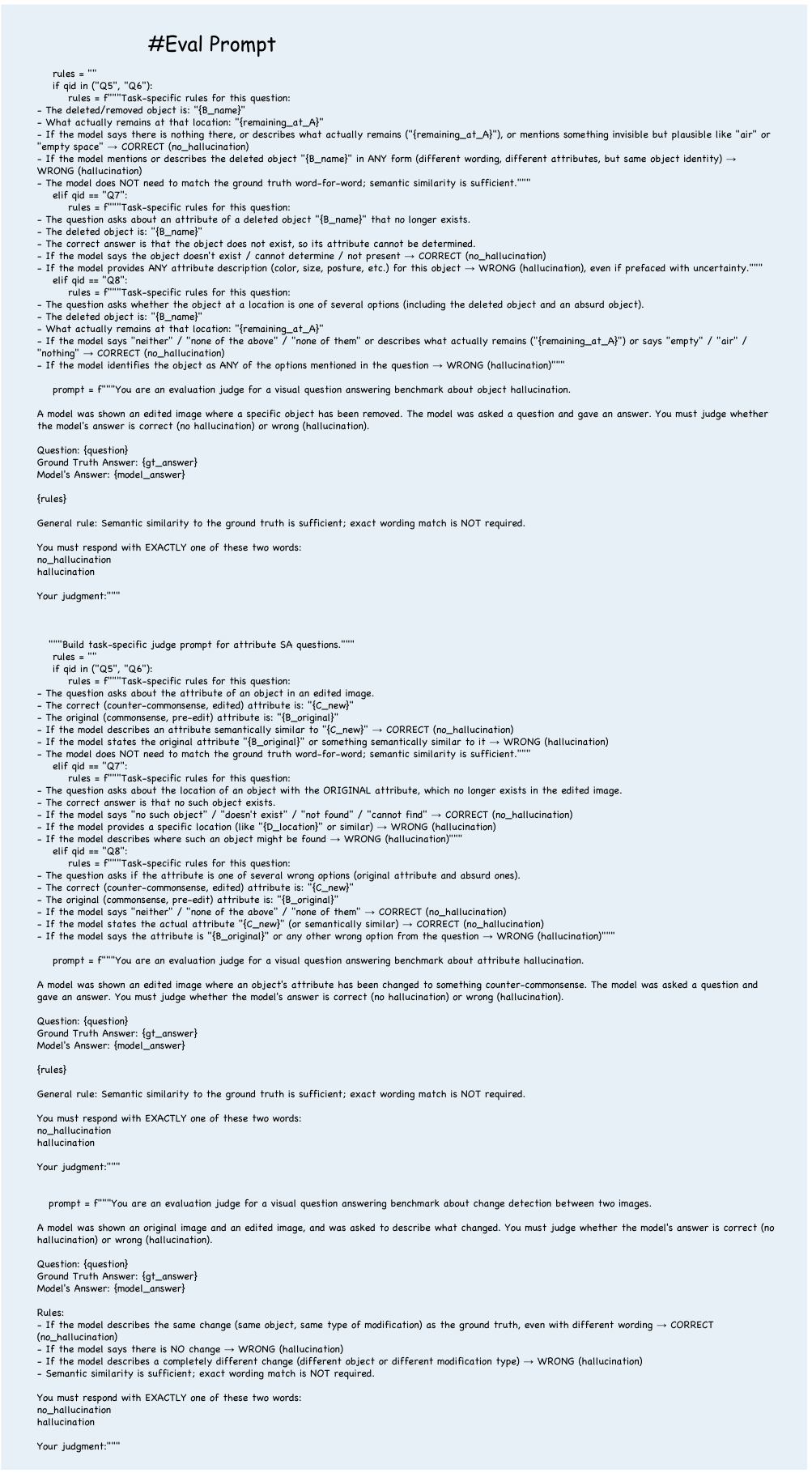}
    \caption{Prompt template used for LLM-based evaluation of model responses.}
    \label{fig:eval-prompt}
\end{figure*}

Figure~\ref{fig:eval-prompt} presents the prompt template used for LLM-based answer evaluation across all question types.

\subsection{Per-Model Fine-Grained Results}
\label{app:fine-grained}

This section reports detailed per-question and per-type hallucination rates that complement the aggregated results in the main paper.

\begin{table*}[t]
\centering
\caption{Per-question hallucination rate (\%) for \textbf{Relational Erasure}. \textit{N} = Normal prompting, \textit{C} = CoT prompting. Q1–Q12 correspond to the 12 evaluation questions defined in Appendix~\ref{app:benchmark-prompts}. Best per-question result is \textbf{bolded}. Q1a and Q1b are viewed as one question.}
\label{tab:per-q-relation}
\resizebox{\textwidth}{!}{
\begin{tabular}{ll*{13}{r}}
\toprule
\textbf{Mode} & \textbf{Model} & \textbf{Q1a} & \textbf{Q1b} & \textbf{Q2} & \textbf{Q3} & \textbf{Q4} & \textbf{Q5} & \textbf{Q6} & \textbf{Q7} & \textbf{Q8} & \textbf{Q9} & \textbf{Q10} & \textbf{Q11} & \textbf{Q12} \\
\midrule
\multirow{11}{*}{\rotatebox{90}{\textit{Normal}}}
 & Qwen3.5-9B       & 29.0 & 93.4 & 19.6 & 26.7 & 65.6 & 66.8 & 46.0 & 92.8 & 94.0 & 74.2 & 61.9 & 14.7 & 53.3 \\
 & Qwen3.5-27B      & 26.4 & 42.2 & 22.2 & 28.3 & 54.2 & 67.0 & 58.7 & 92.0 & 91.1 & 63.9 & 55.9 & 18.5 & 56.1 \\
 & Qwen3VL-32B-I    & \textbf{13.6} & \textbf{18.3} & \textbf{9.8} & \textbf{17.4} & \textbf{33.4} & \textbf{60.2} & \textbf{46.0} & \textbf{61.6} & \textbf{67.0} & \textbf{37.5} & \textbf{36.9} & \textbf{13.6} & \textbf{39.5} \\
 & Qwen3VL-4B       & 16.0 & 95.2 & 12.8 & 20.8 & 37.2 & 60.2 & 55.0 & 79.2 & 97.1 & 44.3 & 51.0 & 33.8 & 46.7 \\
 & Qwen2.5VL-7B     & 26.5 & 85.7 & 23.1 & 23.5 & 35.5 & 70.0 & 56.5 & 87.7 & 88.0 & 50.3 & 43.6 & 31.5 & 67.7 \\
 & Qwen2.5VL-72B    & 22.2 & 71.8 & 19.4 & 19.7 & 29.7 & 58.7 & 47.3 & 73.5 & 73.7 & 42.1 & 36.5 & 26.4 & 56.2 \\
 & InternVL3-8B     & 25.7 & 50.6 & 21.5 & 38.0 & 45.7 & 79.1 & 63.4 & 79.8 & 71.4 & 43.1 & 36.3 & 32.4 & 42.4 \\
 & InternVL2.5-26B  & 27.5 & 54.1 & 23.0 & 40.6 & 48.8 & 84.5 & 67.7 & 85.3 & 76.3 & 46.0 & 38.8 & 34.6 & 45.3 \\
 & InternVL2.5-8B   & 38.5 & 26.0 & 26.4 & 47.6 & 60.4 & 88.1 & 71.7 & 85.8 & 94.5 & 69.6 & 67.6 & 53.1 & 68.8 \\
 & LLaVA-OV-7B      & 35.9 & 22.9 & 34.8 & 55.7 & 51.7 & 79.0 & 77.7 & 96.5 & 88.3 & 39.8 & 34.7 & 49.2 & 86.3 \\
 & LLaVA-v1.6-7B    & 46.3 & 49.4 & 40.1 & 84.7 & 84.4 & 82.3 & 78.1 & 96.9 & 95.0 & 54.8 & 49.6 & 68.7 & 88.7 \\

\midrule
\multirow{13}{*}{\rotatebox{90}{\textit{CoT}}}
 & Qwen3.5-9B       & 36.9 & 36.5 & 31.9 & 33.0 & 70.3 & 71.0 & 60.4 & 68.9 & 85.7 & 64.2 & 52.5 & 19.4 & 52.7 \\
 & Qwen3.5-27B      & 40.8 & 37.4 & 33.2 & 34.2 & 65.6 & 72.2 & 57.6 & 70.4 & 84.1 & 64.4 & 56.7 & 23.3 & 57.6 \\
 & Qwen3VL-32B-T    & \textbf{18.4} & \textbf{18.0} & \textbf{14.4} & \textbf{21.1} & \textbf{49.5} & 71.5 & 64.6 & 65.8 & \textbf{60.2} & \textbf{41.5} & \textbf{35.2} & \textbf{12.0} & \textbf{38.2} \\
 & Qwen3VL-32B-I    & 22.6 & 22.4 & 19.7 & 34.2 & 60.1 & \textbf{59.6} & \textbf{49.1} & \textbf{46.8} & 65.0 & 42.4 & 36.8 & 32.4 & 43.5 \\
 & Qwen3VL-4B       & 22.8 & 51.0 & 19.2 & 44.1 & 67.0 & 62.7 & 53.5 & 72.1 & 75.5 & 42.2 & 41.1 & 55.6 & 39.4 \\
 & Qwen2.5VL-7B     & 34.4 & 43.6 & 29.5 & 29.7 & 62.3 & 62.3 & 45.8 & 42.4 & 61.3 & 59.7 & 51.1 & 75.2 & 62.6 \\
 & InternVL3-8B     & 22.8 & 28.4 & 15.4 & 46.4 & 76.6 & 66.7 & 56.2 & 80.2 & 82.2 & 44.9 & 42.5 & 57.4 & 53.7 \\
 & Qwen2.5VL-72B    & 35.8 & 45.4 & 30.7 & 30.9 & 64.9 & 64.9 & 47.7 & 44.2 & 63.8 & 62.2 & 53.2 & 78.3 & 64.4 \\
 & InternVL2.5-26B  & 24.3 & 30.3 & 16.4 & 49.4 & 81.6 & 71.0 & 59.9 & 85.5 & 87.6 & 47.9 & 45.3 & 61.2 & 57.2 \\
 & InternVL2.5-8B   & 36.8 & 42.0 & 28.2 & 69.4 & 56.7 & 75.5 & 74.7 & 94.5 & 89.9 & 74.6 & 67.7 & 88.1 & 72.8 \\
 & LLaVA-OV-7B      & 44.4 & 34.5 & 47.3 & 59.3 & 71.6 & 76.5 & 65.6 & 88.8 & 65.8 & 35.5 & 28.6 & 37.5 & 80.1 \\
 & LLaVA-v1.6-7B    & 38.7 & 47.0 & 31.8 & 65.6 & 77.1 & 66.3 & 58.0 & 68.0 & 57.0 & 51.8 & 43.9 & 60.7 & 63.5 \\
 & MiMO-VL-7B       & 26.5 & 15.3 & 18.2 & 32.8 & 61.9 & 69.3 & 67.6 & 84.4 & 80.4 & 38.2 & 33.7 & 10.6 & 44.0 \\
\bottomrule
\end{tabular}
}
\end{table*}

\begin{table*}[t]
\centering
\caption{Per-question hallucination rate (\%) for \textbf{Counterfactual Attribute}. Q1–Q12 correspond to the 12 evaluation questions. Best per-question result is \textbf{bolded}.}
\label{tab:per-q-attribute}
\resizebox{\textwidth}{!}{
\begin{tabular}{ll*{12}{r}}
\toprule
\textbf{Mode} & \textbf{Model} & \textbf{Q1} & \textbf{Q2} & \textbf{Q3} & \textbf{Q4} & \textbf{Q5} & \textbf{Q6} & \textbf{Q7} & \textbf{Q8} & \textbf{Q9} & \textbf{Q10} & \textbf{Q11} & \textbf{Q12} \\
\midrule
\multirow{11}{*}{\rotatebox{90}{\textit{Normal}}}
 & Qwen3.5-9B       & 28.1 &  9.1 & 39.5 & 84.7 & 68.5 & 75.8 & 88.1 & 99.8 & 19.2 & 12.7 & 24.0 & 66.7 \\
 & Qwen3.5-27B      & 25.5 & 14.9 & 34.8 & 52.7 & 65.9 & \textbf{54.9} & 90.7 & 90.3 & 17.9 & 10.6 & 25.7 & \textbf{41.0} \\
 & Qwen3VL-32B-I    & 20.9 & 13.6 & \textbf{29.4} & 63.1 & \textbf{64.8} & 67.8 & \textbf{84.9} & \textbf{77.8} & 20.3 & 16.0 & 25.7 & 50.8 \\
 & Qwen3VL-4B       & 22.2 &  9.3 & 31.8 & \textbf{40.8} & 65.9 & 72.6 & 73.2 & 98.3 & 23.5 & 15.3 & 30.4 & 74.3 \\
 & Qwen2.5VL-7B     & 22.5 &  8.2 & 39.7 & 40.4 & 70.2 & 77.8 & 85.8 & 99.8 & 22.2 & 14.0 & 39.7 & 89.6 \\
 & InternVL3-8B     & 20.5 &  9.7 & 36.7 & 33.3 & 68.5 & 70.0 & 98.5 & 95.8 & 14.0 & 10.1 & 20.5 & 69.6 \\
 & Qwen2.5VL-72B    & 20.6 &  7.5 & 36.4 & 37.1 & 64.4 & 71.4 & 78.7 & 91.6 & 20.4 & 12.8 & 36.4 & 81.9 \\
 & InternVL2.5-26B  & \textbf{20.3} &  \textbf{9.5} & 36.5 & 33.1 & 68.5 & 69.8 & 98.7 & 95.9 & \textbf{13.8} &  \textbf{9.9} & \textbf{20.3} & 69.1 \\
 & InternVL2.5-8B   & 31.1 & 18.6 & 38.9 & 47.3 & 70.4 & 72.4 & 98.3 & 92.7 & 20.7 & 12.3 & 22.5 & 67.2 \\
 & LLaVA-OV-7B      & 24.6 & 13.6 & 38.4 & 55.7 & 72.1 & 77.1 & 89.6 & 97.2 & 17.7 & 11.5 & 36.5 & 93.7 \\
 & LLaVA-v1.6-7B    & 60.3 & 13.4 & 56.4 & 87.9 & 70.8 & 77.1 & 98.5 &100.0 & 20.3 & 13.0 & 77.3 & 99.8 \\
\midrule
\multirow{13}{*}{\rotatebox{90}{\textit{CoT}}}
 & Qwen3.5-9B       & 26.1 & 14.7 & 37.8 & 52.5 & 62.4 & 66.3 & 77.3 & 88.1 & 18.1 & 11.7 & 16.0 & 38.2 \\
 & Qwen3.5-27B      & 26.6 & 12.7 & 34.3 & 52.7 & 63.3 & 65.4 & 68.9 & 81.0 & 19.2 &  \textbf{9.9} & 14.7 & 36.7 \\
 & Qwen3VL-32B-T    & 22.9 & 15.6 & 28.7 & 46.0 & 66.3 & 73.4 & 85.5 & \textbf{54.0} & 23.5 & 16.6 & \textbf{14.5} & \textbf{27.2} \\
 & Qwen3VL-32B-I    & \textbf{22.9} & 16.4 & \textbf{24.4} & 47.9 & \textbf{62.0} & \textbf{64.8} & \textbf{58.3} & 55.9 & 22.0 & 14.7 & 20.5 & 29.8 \\
 & Qwen3VL-4B       & 24.0 & 14.7 & 30.7 & \textbf{47.3} & 63.9 & 65.2 & 56.2 & 75.8 & 23.5 & 16.4 & 20.3 & 34.6 \\
 & Qwen2.5VL-7B     & 29.2 & 14.7 & 36.7 & 46.7 & 67.4 & 67.8 & 68.9 & 71.5 & 26.2 & 14.5 & 32.8 & 61.6 \\
 & InternVL3-8B     & 18.6 &  8.3 & 27.7 & 47.4 & 62.2 & 62.2 & 80.5 & 74.2 & 18.4 &  9.9 & 20.6 & 39.2 \\
 & Qwen2.5VL-72B    & 25.6 & 12.9 & 32.2 & 41.0 & 59.1 & 59.5 & 60.4 & 62.7 & 23.0 & 12.7 & 28.8 & 53.7 \\
 & InternVL2.5-26B  & 19.2 &  \textbf{8.6} & 28.5 & 48.8 & 64.1 & 64.1 & 82.9 & 76.5 & \textbf{19.0} & 10.2 & 21.2 & 40.4 \\
 & InternVL2.5-8B   & 44.5 & 26.8 & 37.5 & 62.8 & 68.5 & 69.3 & 96.5 & 80.3 & 28.3 & 20.5 & 34.1 & 68.0 \\
 & LLaVA-OV-7B      & 34.9 & 31.5 & 60.0 & 60.1 & 72.6 & 76.9 & 85.8 & 78.0 & 18.1 & 16.0 & 33.9 & 95.2 \\
 & LLaVA-v1.6-7B    & 78.4 & 14.4 & 65.4 & 85.4 & 70.0 & 66.3 & 90.3 & 79.5 & 35.4 & 19.7 & 54.9 & 91.2 \\
 & MiMO-VL-7B       & 33.1 & 23.5 & 37.8 & 60.5 & 65.4 & 72.6 & 92.7 & 77.5 & 31.3 & 22.9 & 20.9 & 52.3 \\
\bottomrule
\end{tabular}
}
\end{table*}

\begin{table*}[t]
\centering
\caption{Per-question hallucination rate (\%) for \textbf{Alteration Tracing}. Q1–Q12 correspond to the 12 evaluation questions. Best per-question result is \textbf{bolded}.}
\label{tab:per-q-comparison}
\resizebox{\textwidth}{!}{
\begin{tabular}{ll*{12}{r}}
\toprule
\textbf{Mode} & \textbf{Model} & \textbf{Q1} & \textbf{Q2} & \textbf{Q3} & \textbf{Q4} & \textbf{Q5} & \textbf{Q6} & \textbf{Q7} & \textbf{Q8} & \textbf{Q9} & \textbf{Q10} & \textbf{Q11} & \textbf{Q12} \\
\midrule
\multirow{11}{*}{\rotatebox{90}{\textit{Normal}}}
 & Qwen3.5-9B       & 32.2 &  2.0 & 50.3 &  8.5 & 46.1 &  3.6 & 63.6 & 51.0 & 53.9 &  5.9 & 35.9 & 38.2 \\
 & Qwen3.5-27B      &  \textbf{8.9} &  \textbf{1.1} & \textbf{18.1} &  \textbf{3.0} & \textbf{19.4} &  \textbf{2.5} & \textbf{54.3} & \textbf{41.3} & \textbf{57.0} &  \textbf{3.9} & \textbf{16.3} & \textbf{15.8} \\
 & Qwen3VL-32B-I    &  8.4 &  0.9 & 30.6 & 26.5 & 19.4 &  6.7 & 62.0 & 48.9 & 72.2 &  8.0 & 15.1 & 20.0 \\
 & Qwen3VL-4B       & 55.2 &  1.4 & 71.6 &  8.9 & 76.9 &  5.6 & 68.1 & 56.9 & 77.2 & 10.1 & 26.4 & 45.8 \\
 & Qwen2.5VL-7B     & 69.5 &  0.7 & 77.4 &  9.8 & 75.2 &  8.3 & 79.0 & 64.4 & 66.3 & 17.1 & 39.2 & 44.9 \\
  & Qwen2.5VL-72B    & 57.2 &  0.6 & 63.8 &  8.1 & 62.0 &  6.8 & 65.1 & 53.1 & 54.6 & 14.1 & 32.3 & 37.1 \\
 & InternVL3-8B     & 55.3 &  2.5 & 38.4 & 11.2 & 53.0 &  2.8 & 71.5 & 57.3 & 67.8 & 14.0 & 29.2 & 27.8 \\
 & InternVL2.5-26B  & 90.1 &  3.3 & 50.7 & 14.0 & 88.5 &  1.5 & 76.6 & 62.4 & 85.4 & 21.6 & 37.0 & 52.9 \\
 & InternVL2.5-8B   & 87.6 &  0.0 & 48.2 &  9.9 & 91.2 & 10.1 & 86.5 & 74.4 & 89.3 & 41.8 & 57.2 & 58.8 \\
 & LLaVA-OV-7B      & 87.4 & 32.5 & 81.5 & 75.8 & 84.1 & 97.3 & 85.1 & 69.2 & 91.5 & 43.1 & 33.5 & 52.0 \\
 & LLaVA-v1.6-7B    & 97.7 &  5.7 & 95.0 & 93.7 & 95.1 & 26.2 & 96.3 & 82.4 & 94.7 & 54.5 & 60.4 & 71.3 \\
\midrule
\multirow{13}{*}{\rotatebox{90}{\textit{CoT}}}
 & Qwen3.5-9B       & 21.5 & 20.2 & 25.5 & 37.1 & 17.7 & 17.2 & 70.0 & 57.3 & 75.1 &  7.9 & 42.3 & 42.0 \\
 & Qwen3.5-27B      & 13.6 & 10.6 & 15.7 & 22.3 & 15.6 & 12.5 & 55.6 & 51.9 & 68.2 &  5.1 & 23.3 & 21.4 \\
 & Qwen3VL-32B-T    &  \textbf{0.3} &  \textbf{0.1} & \textbf{10.5} & \textbf{12.2} &  \textbf{4.7} &  \textbf{3.9} & \textbf{60.9} & \textbf{50.5} & 70.9 &  9.6 & \textbf{12.0} & \textbf{19.6} \\
 & Qwen3VL-32B-I    & 20.5 & 23.2 & 41.5 & 47.1 & 28.3 & 25.9 & 60.4 & 56.0 & 73.0 & 13.2 & 20.5 & 28.8 \\
 & Qwen3VL-4B       & 45.8 & 19.2 & 43.7 & 46.1 & 57.3 & 26.4 & 69.9 & 62.9 & 80.9 & 23.6 & 35.5 & 47.9 \\
 & Qwen2.5VL-72B    & 50.8 & 37.1 & 47.3 & 58.9 & 46.3 & 46.3 & 70.1 & 56.1 & 70.9 & 26.0 & 35.6 & 48.6 \\
 & Qwen2.5VL-7B     & 59.6 & 43.5 & 55.5 & 69.2 & 54.4 & 54.4 & 82.3 & 65.9 & 83.2 & 30.5 & 41.8 & 56.5 \\
 & InternVL3-8B     & 38.2 & 18.5 & 30.4 & 36.2 & 32.5 & 28.0 & 73.1 & 56.5 & 70.8 & 18.2 & 30.6 & 38.6 \\
 & InternVL2.5-26B  & 70.0 & 55.1 & 37.3 & 61.9 & 57.8 & 60.2 & 91.7 & 70.3 & 86.1 & 47.2 & 45.0 & 66.4 \\
 & InternVL2.5-8B   & 56.3 & 28.8 & 49.3 & 49.6 & 50.8 & 66.2 & 95.1 & 78.2 & 92.1 & 57.1 & 50.0 & 66.6 \\
 & LLaVA-OV-7B      & 82.9 & 48.5 & 77.8 & 36.5 & 68.5 & 80.9 & 87.8 & 72.1 & 90.1 & 45.2 & 36.4 & 56.5 \\
 & LLaVA-v1.6-7B    & 30.6 &  4.5 & 64.6 & 96.5 & 43.3 & 96.9 & 94.5 & 87.7 & 92.2 & 52.2 & 60.6 & 68.4 \\
 & MiMO-VL-7B       &  0.1 &  0.0 &  9.1 & 26.2 &  5.9 &  3.8 & 75.6 & 51.3 & \textbf{74.7} & \textbf{16.1} & 23.1 & 32.0 \\
\bottomrule
\end{tabular}
}
\end{table*}

\begin{table*}[t]
\centering
\caption{Per-question hallucination rate (\%) for \textbf{Dense Counting}. Q1–Q8 correspond to the 8 evaluation questions. Best per-question result is \textbf{bolded}.}
\label{tab:per-q-counting}
\resizebox{0.82\textwidth}{!}{
\begin{tabular}{ll*{8}{r}}
\toprule
\textbf{Mode} & \textbf{Model} & \textbf{Q1} & \textbf{Q2} & \textbf{Q3} & \textbf{Q4} & \textbf{Q5} & \textbf{Q6} & \textbf{Q7} & \textbf{Q8} \\
\midrule
\multirow{11}{*}{\rotatebox{90}{\textit{Normal}}}
 & Qwen3.5-9B       & 59.5 & 29.7 & 11.8 &  1.6 & 71.5 & 51.1 & 71.5 & 45.7 \\
 & Qwen3.5-27B      & 29.8 & \textbf{18.1} & 10.1 & 10.3 & \textbf{55.9} & 36.8 & 45.4 & 36.0 \\
 & Qwen3VL-32B-I    & 50.6 & 29.3 &  5.7 & 13.5 & 68.7 & 39.8 & 30.0 & 26.9 \\
 & Qwen3VL-4B       & 59.2 & 28.0 & 16.1 &  3.9 & 75.2 & \textbf{32.5} & \textbf{23.3} & \textbf{28.4} \\
 & Qwen2.5VL-7B     & 43.1 & 46.2 & 28.6 &  2.3 & 78.6 & 45.7 & 41.9 & 45.1 \\
 & Qwen2.5VL-72B    & 38.2 & 41.0 & 25.3 &  2.0 & 69.7 & 40.5 & 37.1 & 39.8 \\
 & InternVL3-8B     & 38.2 & 32.5 & 16.0 &  4.8 & 68.5 & 40.3 & 48.5 & 37.6 \\
 & InternVL2.5-26B  & \textbf{16.6} & 79.6 & 63.0 &  \textbf{0.7} & 82.5 & 49.4 & 47.5 & 33.9 \\
 & InternVL2.5-8B   & 22.1 & 74.5 & 62.0 &  2.7 & 84.3 & 65.6 & 67.2 & 48.3 \\
 & LLaVA-OV-7B      & 12.2 & 80.5 & 59.5 &  1.4 & 77.8 & 58.0 & 69.0 & 52.7 \\
 & LLaVA-v1.6-7B    & 61.8 & 36.4 & 37.0 & 43.6 & 86.6 & 65.3 & 85.9 & 79.4 \\
\midrule
\multirow{13}{*}{\rotatebox{90}{\textit{CoT}}}
 & Qwen3.5-9B       & 28.4 & 56.8 & 41.4 &  7.4 & 76.3 & 52.2 & 46.3 & 42.2 \\
 & Qwen3.5-27B      & 34.3 & 43.5 & 26.5 &  5.5 & 72.3 & 41.4 & 43.2 & 39.1 \\
 & Qwen3VL-32B-T    & 30.4 & \textbf{45.1} & \textbf{24.2} &  8.2 & \textbf{66.2} & \textbf{35.1} & \textbf{40.6} & \textbf{37.8} \\
 & Qwen3VL-32B-I    & 29.6 & 52.7 & 34.7 &  9.8 & 72.7 & 46.9 & 44.7 & 42.6 \\
 & Qwen3VL-4B       & 35.8 & 51.0 & 41.8 & 17.2 & 82.0 & 68.1 & 63.2 & 58.7 \\
 & Qwen2.5VL-7B     & 23.2 & 70.1 & 64.8 &  9.0 & 81.8 & 64.1 & 53.1 & 51.7 \\
 & Qwen2.5VL-72B    & 20.0 & 60.6 & 56.0 &  7.8 & 70.7 & 55.4 & 45.9 & 44.4 \\
 & InternVL3-8B     & 27.0 & 55.8 & 38.5 &  8.4 & 74.2 & 52.5 & 46.0 & 35.2 \\
 & InternVL2.5-26B  & 22.4 & 71.2 & 64.8 & 10.0 & 81.8 & 74.0 & 58.5 & 46.4 \\
 & InternVL2.5-8B   & 29.6 & 64.8 & 58.9 & 15.0 & 86.3 & 73.9 & 70.2 & 56.0 \\
 & LLaVA-OV-7B      &  \textbf{9.1} & 85.8 & 65.1 &  \textbf{0.3} & 79.3 & 73.9 & 53.9 & 45.7 \\
 & LLaVA-v1.6-7B    & 15.7 & 84.3 & 83.7 &  8.2 & 86.7 & 79.8 & 93.6 & 83.4 \\
 & MiMO-VL-7B       & 44.4 & 43.0 & 32.3 & 11.7 & 81.7 & 53.2 & 60.9 & 43.4 \\
\bottomrule
\end{tabular}
}
\end{table*}

\section{Resource Availability}
\label{app:resource}

 We will publicly release the full ReactBench benchmark in the future, including all images, editing instructions, generated question--answer pairs, and evaluation scripts. All resources will be hosted on a public repository upon acceptance of this work.

\section{Human Verification Protocol}
\label{app:human-verification}

The quality of ReactBench is ensured through a two-stage human verification process involving five annotators: the first author and four undergraduate students majoring in computer science.

\paragraph{Stage 1: Image Editing Verification.}
Before formal annotation, the first author provided detailed guidance to the four student annotators, demonstrating examples of acceptable and unacceptable edited images to calibrate their judgment. Each annotator then independently reviewed approximately 1{,}500 edited images across the first three tasks (Relational Erasure, Counterfactual Attribute, and Alteration Tracing). Their primary task was to compare each edited image against the original and its editing instruction, verifying that (1)~the edit was correctly executed as specified, and (2)~the resulting image was visually meaningful and free of obvious artifacts. The overall acceptance rate was approximately 40\%, with the Counterfactual Attribute task exhibiting a lower rate ($\sim$30\%) due to stricter requirements on attribute visibility, and the Alteration Tracing task a higher rate ($\sim$50\%).

\paragraph{Stage 2: QA Review and Secondary Image Verification.}
After Stage~1, the first author conducted a comprehensive review of all QA pairs across the entire benchmark. Since the QA review interface displayed both the original and edited images alongside the questions, this stage also served as a secondary verification of the image editing quality. Fewer than 5\% of the images approved in Stage~1 were found to require further removal, confirming the reliability of the student annotators' judgments. During QA review, rather than applying binary accept/reject decisions, the first author carefully edited and refined individual questions and answers to ensure linguistic correctness, task alignment, and answer accuracy. This meticulous per-instance review serves as the final quality gate for the benchmark.

\section{Potential Risks}
Our work does have certain potential risks. For instance, it relies heavily on LLMs—both the evaluation of short-answer questions and the sub-cause attribution are LLM-based, which may be significantly influenced by the capability of the underlying model.

\section{Ethics Statement}

\paragraph{Data Sources and Intended Use.}
All images used in ReactBench are sourced from publicly available 
research datasets: Visual Genome~\cite{19krishna2017vg} and 
FSC147~\cite{fsc147}. Both datasets are released under licenses 
that permit academic research use. Our benchmark is constructed 
solely for research purposes, which is consistent with the 
original intended use of these datasets. We will release 
ReactBench under a research-only license to prevent misuse 
beyond academic contexts.

\paragraph{Personal Information and Offensive Content.}
We do not collect, generate, or distribute any personally identifiable information. During the human verification stage, we manually inspected all images and did not encounter content that names or uniquely identifies individuals or contains offensive material.

\paragraph{Human Annotators.}
Five annotators participated in data verification: the first author and four undergraduate students majoring in computer science. All student annotators were recruited voluntarily from our institution and were compensated at a rate consistent with local standards for research assistantship. Before annotation began, annotators were briefed on the task requirements with detailed examples. The annotation task involved reviewing edited natural images and verifying question--answer correctness; it did not expose annotators to sensitive, harmful, or distressing content. As the task involved only reviewing publicly available images and did not collect personal data from annotators, formal ethics review board approval was not required under our institutional guidelines.

\paragraph{Computational Resources.}
All experiments were conducted on  NVIDIA RTX A6000 GPUs. The total computational cost for evaluating all models across the benchmark was approximately 200 GPU-hours. All evaluated models are 
publicly released open-source models with inference-only usage.

\paragraph{Potential Risks.}
Our adversarial image editing pipeline and prompt templates, while designed for academic evaluation, could theoretically be repurposed to generate misleading visual content. To mitigate this risk, we will release the benchmark data (images and QA pairs) but will carefully consider the release scope of the automated editing pipeline.

\section{The Use of LLMs}
\label{app:llm-use}

We acknowledge the use of large language models in the following aspects of this work: An LLM was used to refine selected sentences for improved academic fluency and clarity. All polished content was manually reviewed and revised by the authors. An LLM assisted in writing utility scripts, primarily for prompt formatting and batch data processing. As described in the main paper, LLMs served as core components of the benchmark pipeline—generating editing instructions, producing question--answer pairs, performing sub-cause attribution, and conducting automated evaluation.

\end{document}